\documentclass{article}

 \usepackage[preprint]{neurips_2026}

\usepackage[utf8]{inputenc} 
\usepackage[T1]{fontenc}    
\usepackage{hyperref}       
\usepackage{url}            
\usepackage{booktabs}       
\usepackage{amsfonts}       
\usepackage{nicefrac}       
\usepackage{microtype}      

\usepackage[table]{xcolor}
\usepackage{amsmath}
\usepackage{xspace} 
\usepackage{bm}
\newcommand{\ourbench}{Re$^2$Math\xspace}
\AtBeginDocument{%
  }
\usepackage{tcolorbox}
\tcbuselibrary{breakable} 
\usepackage{booktabs}
\usepackage{tabularx}
\usepackage{array}
\usepackage{ragged2e}
\usepackage{makecell}
\usepackage{xurl} 
\usepackage{amssymb}

\newcolumntype{P}[1]{>{\RaggedRight\arraybackslash}p{#1}}
\newcommand{\caseid}[1]{{\footnotesize\path{#1}}}
\title{Re$^2$Math: Benchmarking Theorem Retrieval in Research-Level Mathematics}

\author{%
  Zicheng Lyu\thanks{Equal contribution} \\
  Fudan University\\
  Shanghai, China \\
  \texttt{lyuzicheng@gmail.com} \\
  \And
  Wenjie Yang\footnotemark[1] \\
  Fudan University\\
  Shanghai, China \\
  \texttt{yangwj24@m.fudan.edu.cn}\\
  \AND
  Shengzhong Zhang \\
  Nanjing University of Aeronautics and Astronautics\\
  Nanjing, Jiangsu, China \\
  \texttt{szzhang@nuaa.edu.cn}\\
  \And
  Zengfeng Huang\thanks{Corresponding author.} \\
  Fudan University\\
  Shanghai Innovation Institute\\
  Shanghai, China \\
  \texttt{huangzf@fudan.edu.cn}
}

\begin{document}

\maketitle

\begin{abstract}
Large language models are increasingly capable at closed-world mathematical reasoning, but research assistance also requires source-grounded use of the literature. When a proof reaches a non-trivial step, a useful assistant should determine whether the needed tool (e.g., a lemma) already exists, identify a suitable scholarly source, and verify that its assumptions align with the current proof context.
To rigorously evaluate such capabilities, we introduce \ourbench, a benchmark for \emph{tool-grounded retrieval} from partial mathematical proofs. Each instance is built from a candidate instrumental citation in the proof of a main theorem, with hierarchical context and an optional leakage-controlled anchor hint. We also make the task source-grounded yet citation-agnostic in that any admissible theorem sufficient for the proof transition is accepted.
Evaluation uses a release-frozen retrieval artifact, ensuring reproducibility, while the benchmark itself supports automatic, continual expansion with newly constructed instances. On the current benchmark test set, the best fixed-judge ToolAcc reaches 7.0\%, despite substantially higher rates of source grounding, indicating that current systems often retrieve valid statements but fail to establish their applicability to the local proof step.
By decoupling citation recall, grounding, and proof-gap sufficiency, \ourbench{} transforms literature-grounded mathematical tool use into a controlled diagnostic task.
\end{abstract}

\section{Introduction}
\label{sec:intro}

Large language models (LLMs) have improved rapidly on mathematical reasoning benchmarks, from grade-school arithmetic to olympiad-style problem solving \cite{cobbe2021training,hendrycks2021measuring,he2024olympiadbench,gao2024omni,sun2025challenging,glazer2024frontiermath}. Yet a useful mathematical research assistant must do more than solve self-contained problems. Research proofs often reach steps that require an external theorem, estimate, compactness criterion, representation result, or gluing principle. At such points, the assistant should locate prior work, identify an admissible source, and check whether the source's assumptions apply to the current proof state.

This motivates \emph{tool-grounded retrieval}. A model may state a theorem-like claim from memory, but a claim without provenance leaves open where the result was established, how it should be attributed, and whether its hypotheses match the proof at hand. The object to be evaluated is therefore a source--tool pair: a source that supports the proposed mathematical tool, and a statement whose assumptions and conclusion are sufficient for the local proof transition. This is different from both unsupported theorem suggestion and exact citation recovery.

Existing benchmarks only partially capture this capability. Outcome-oriented math datasets ask for a final answer or complete solution from a largely self-contained prompt, while formal proving benchmarks evaluate search inside fixed formal libraries \cite{zheng2021minif2f,azerbayev2023proofnet,tsoukalas2024putnambench}. They do not isolate open-world literature use from partial research-proof context, nor do they test whether a model can turn a proof gap into a verifiable source--tool pair.

We introduce \ourbench, a benchmark for tool-grounded retrieval. We derive proof gaps
from citations that our construction pipeline identifies as instrumental: citations that appear to
supply an external mathematical tool for a local transition inside a main-theorem proof. Each instance
supplies hierarchical context---a global setup and a local proof window immediately preceding the
citation---and an assisted track additionally provides a sanitized anchor hint. Given this context, an
agent must infer the proof need, generate search queries, select a source, and extract or faithfully
restate a theorem-like statement. Appendix~\ref{app:construction} audits this construction signal
and reports the remaining risks in instrumentality and anchor sanitization.

The primary leaderboard uses controlled, model-authored retrieval. Models may propose their own search queries, but query execution is performed through a shared release-frozen artifact rather than through model-native web-search, browser, file-reading, or deep-research modes. This restriction does not remove retrieval from the task; it fixes the information-access boundary so that hidden ranking, browsing, source-opening, and context-packaging differences do not dominate the mathematical source--tool policy being evaluated. \ourbench{} therefore asks whether a system can produce a verifiable scholarly justification under a frozen release, not whether it can reproduce the original author's historical search process. Appendix~\ref{app:deep_research_exclusion} details this benchmark contract.

A central design choice is that \ourbench{} is source-grounded but citation-invariant. The author-cited paper is useful for construction and diagnostics, but it is not the definition of correctness. A system receives credit whenever it returns any admissible source-grounded statement that closes the gap. This matches research practice, where the same tool may appear in multiple papers, books, surveys, or stable preprints, and a stronger alternative theorem may suffice if its assumptions are satisfied.

\ourbench{} is built by an automated published-first, arXiv-backed pipeline and released as a versioned snapshot of a living benchmark. We evaluate source-grounded mathematical tool retrieval under a fixed-budget, shared-backend scholarly-search setting. In this setting, the best evaluated model on the benchmark test set (Eval-200) reaches only \(7.0\%\) \(\mathrm{ToolAcc}\), despite substantially higher grounding rates. We treat the leaderboard as a diagnostic rather than a ranking endpoint: the main empirical signal is that current systems often find or restate real mathematical claims without making them usable for the exact proof transition.

Our contributions are summarized as follows:
\begin{itemize}
    \item \textbf{Task formulation.} We formalize tool-grounded retrieval from partial research proofs with hierarchical context, raw and assisted tracks, and a citation-invariant success criterion.
    \item \textbf{Living benchmark.} We construct a versioned benchmark from instrumental proof citations via a published-first, arXiv-backed pipeline that \emph{automatically} incorporates new instances as the literature grows, while keeping paper-facing evaluation fixed to a frozen held-out release.
    \item \textbf{Diagnostic evaluation.} We provide a shared-backend protocol with \emph{ToolAcc} as the primary metric and diagnostics that separate citation-side retrieval, source grounding, proof-gap sufficiency, oracle-source behavior, and alternative-source success.
\end{itemize}

\section{Related Work}
\label{sec:related}

\paragraph{Mathematical reasoning and proof benchmarks.}
GSM8K and MATH established standard benchmarks for multi-step mathematical reasoning \cite{cobbe2021training,hendrycks2021measuring}; later datasets pushed toward harder, refreshed, or leakage-resistant evaluation, including OlympiadBench, Omni-MATH, OlymMATH, FrontierMath, HARP, LiveBench, and LiveAoPSBench \cite{he2024olympiadbench,gao2024omni,sun2025challenging,glazer2024frontiermath,yue2024harp,white2024livebench,mahdavi2025leveraging}. Formal benchmarks such as miniF2F, ProofNet, and PutnamBench evaluate proof production in mechanically checkable settings \cite{zheng2021minif2f,azerbayev2023proofnet,tsoukalas2024putnambench}. These benchmarks measure closed-world problem solving, proof search, or formalization. \ourbench\ targets a different research-assistance primitive: from a partial research proof, a system must identify an external mathematical tool, ground it in an admissible scholarly source, and check that it suffices for the local proof transition.

\paragraph{Inference-time reasoning and controlled retrieval.}
Modern reasoning systems increasingly rely on inference-time computation and verification, from Chain-of-Thought, self-consistency, and Tree of Thoughts to frontier and verifier-guided mathematical systems \cite{wei2022chain,wang2022self,yao2023tree,openai2024o1,guo2025deepseek,wang2024math,xin2024deepseek,hubert2025olympiad}. Retrieval and attribution benchmarks such as KILT, BEIR, and ALCE study shared retrieval settings, external-memory access, and citation-supported generation \cite{petroni2021kilt,thakur2021beir,gao2023alce}. \ourbench\ is complementary: rather than proposing a new reasoning algorithm or a general RAG benchmark, it fixes a controlled retrieval interface and evaluates the mathematical policy mapping a proof state to a grounded source--tool pair.

\paragraph{Scientific literature use and citation recommendation.}
Scientific claim-verification benchmarks such as SciFact and SciFact-Open retrieve evidence from papers to support or refute scientific claims \cite{wadden2020scifact,wadden2022scifactopen}. Literature assistants such as PaperQA and OpenScholar search scientific papers to produce citation-backed answers \cite{lala2023paperqa,asai2024openscholar}, while SPECTER and SciRepEval evaluate scientific document representations for search, ranking, and recommendation \cite{cohan2020specter,singh2023scirepeval}. Context-aware citation recommendation retrieves papers from a local writing context \cite{he2010contextaware,farber2020citationrecommendation}. These works connect textual contexts to scholarly sources, but typically target article recommendation, claim evidence, or literature synthesis. In contrast, \ourbench\ uses a mathematical proof state as context and requires a theorem-like tool whose assumptions and conclusion are sufficient for a specific proof gap, not merely topically relevant.

\paragraph{Theorem retrieval and premise selection.}
Retrieval is also central to mathematical reasoning. Natural-language resources such as NaturalProofs and TheoremQA support theorem-conditioned reasoning and theorem use \cite{welleck2021naturalproofs,chen2023theoremqa}; formal premise-selection systems and benchmarks such as MPTP, DeepMath, HolStep, HOList, LeanDojo, LeanSearch, and Magnushammer retrieve accessible premises from a formal library or proof state \cite{urban2006mptp,alemi2016deepmath,kaliszyk2017holstep,bansal2019holist,yang2023leandojo,gao2024semantic,mikula2023magnushammer}. \ourbench{} differs along the evaluation axis: retrieval is mediated by a frozen scholarly-search
interface over an open-world scholarly source space; the output is a sourced theorem-like statement
rather than a premise identifier or relevance score; and success is citation-invariant---any admissible
source receives credit if it grounds a statement sufficient for the proof gap.

\section{Task Definition}
\label{sec:formulation}

We study \emph{tool-grounded retrieval}. Given the proof state immediately before a non-trivial proof transition and access to a scholarly search interface over admissible sources, an agent must return a source--tool pair
\[
(\hat d_i,\hat K_i),
\]
where \(\hat d_i\) is an admissible source and \(\hat K_i\) is a theorem-like statement grounded in that source and sufficient for the transition. The task is \emph{source-grounded} but \emph{citation-invariant}: the selected source must support the returned statement, but it need not be the exact source cited by the original author. The author citation is one human witness for the gap, not the definition of correctness.

This formulation separates \ourbench{} from both unsupported theorem generation and citation classification. A successful system must infer the proof need, express it in searchable language, select a source from noisy candidates, and extract a statement whose assumptions fit the local context. We expose these stages for diagnosis while scoring a single end-to-end object: the returned source--tool pair. A high-level overview of the task is illustrated in Figure~\ref{fig:task_overview}.

\subsection{Instances and Input Tracks}
\label{subsec:hierarchical_context}

Let a proof be segmented into ordered blocks
\[
P=(s_1,\ldots,s_T),
\]
where each block is a sentence, displayed equation, or mathematical environment. Suppose a citation occurrence in block \(s_c\) is treated as a witness for a non-trivial proof transition. We expose only the pre-citation state \(k=c-1\), and write \(\Delta_i\) for the target transition into the citation block.

Each instance \(i\) provides hierarchical context
\[
\mathcal C_i=
(\mathcal C^i_{\mathrm{global}},\mathcal C^i_{\mathrm{local}}(m)),
\qquad
\mathcal C^i_{\mathrm{local}}(m)=
(s_{\max\{1,k-m+1\}},\ldots,s_k).
\]
The global context summarizes the paper setup, assumptions, definitions, and target theorem; the local context is the ordered proof window immediately before the citation. The canonical stored window keeps up to five pre-citation blocks, and local-window diagnostics use \(m\in\{1,3,5\}\).

Each instance may also include a sanitized anchor \(a_i\), a short description of the immediate proof role of the missing tool. The anchor is a planning cue, not a bibliographic shortcut: it should not contain theorem titles, author names, citation keys, DOIs, arXiv identifiers, or source metadata. We define two input tracks:
\[
\mathsf{x}^{i}_{\mathrm{raw}}
:=
(\mathcal C^{i}_{\mathrm{global}},\mathcal C^{i}_{\mathrm{local}}),
\qquad
\mathsf{x}^{i}_{\mathrm{assist}}
:=
(\mathcal C^{i}_{\mathrm{global}},\mathcal C^{i}_{\mathrm{local}},a_i).
\]
The Raw track requires the agent to infer the proof need from context alone. The Assisted track fixes this planning cue so that retrieval, source selection, and extraction can be studied more directly.

For supervision and analysis, each instance stores
\[
\underbrace{\mathsf{x}_i}_{\text{agent input}}
\qquad
\underbrace{\mathsf{y}_i=K_i^\star}_{\text{reference tool witness}}
\qquad
\underbrace{\mathsf{z}_i=d_i^{\mathrm{cite}}}_{\text{citation witness}}.
\]
Here \(K_i^\star\) is one curated theorem-like witness aligned to the gap, and \(d_i^{\mathrm{cite}}\) is the source used by the original author. Together they certify that the transition is externally tool-dependent, but they do not exhaust the valid answer set.

\subsection{Retrieval Interface and Agent Policy}
\label{subsec:agent_policy}

Let \(\mathcal D\) be the admissible scholarly source space and let \(\mathcal K(d)\) denote theorem-like statements stated in, or faithfully entailed by, a source \(d\in\mathcal D\). We model retrieval abstractly by an operator
\[
\rho_N:\mathcal Q^L\to\mathcal D^N,
\]
which maps a bounded query list to a ranked candidate prefix of length at most \(N\). The experimental protocol instantiates \(\rho_N\) with a shared release-frozen retrieval interface; the task definition only assumes that all agents use the same retrieval operator.

An agent proceeds through four conceptual stages:
\[
\mathsf{x}_i
\xrightarrow{\ \pi_{\mathrm{plan}}\ }
\hat a_i
\xrightarrow{\ \pi_{\mathrm{qry}}\ }
Q_i
\xrightarrow{\ \rho_N\ }
\mathcal R_{i,N}
\xrightarrow{\ \pi_{\mathrm{sel}}\ }
\hat d_i
\xrightarrow{\ \pi_{\mathrm{ext}}\ }
\hat K_i .
\]
This diagram is a workflow factorization, not a strict unary composition: selection and extraction may condition on the proof context \(\mathcal C_i\) and the planned proof need \(\hat a_i\). More explicitly,
\[
\hat a_i=
\begin{cases}
\pi_{\mathrm{plan}}(\mathsf{x}^{i}_{\mathrm{raw}}), & \text{Raw track},\\[0.2em]
a_i, & \text{Assisted track},
\end{cases}
\qquad
Q_i=\pi_{\mathrm{qry}}(\mathcal C_i,\hat a_i),
\qquad
\mathcal R_{i,N}=\rho_N(Q_i),
\]
\[
\hat d_i=
\pi_{\mathrm{sel}}(\mathcal C_i,\hat a_i,\mathcal R_{i,N})
\in \mathcal R_{i,N}\cup\{\bot\},
\qquad
\hat K_i=
\pi_{\mathrm{ext}}(\mathcal C_i,\hat a_i,\hat d_i)
\in \mathcal K(\hat d_i)\cup\{\bot\}.
\]
The stages diagnose distinct capabilities: recognizing the missing mathematical role, lexicalizing it for search, choosing a source, and extracting a usable result. Only the final pair
\[
\hat{\mathsf y}_i=(\hat d_i,\hat K_i)
\]
is the official task output.

\subsection{Valid Outputs}
\label{subsec:supervision_validity}

A source--tool pair is valid if the statement is grounded in an admissible source and sufficient for the transition \(\Delta_i\). Admissible sources are scholarly documents external to the local citing proof: the citing paper itself, and its arXiv/published variants, do not count merely by restating the citation context or the proof sentence that generated the instance. They are admissible only if the selected text independently states or proves the external tool being invoked.

We define the valid-output set
\[
\mathcal V_i
:=
\{(d,K)\in\mathcal D\times\mathcal K(d):
\mathsf{Ground}_i(d,K)=1
\ \land\
\mathsf{Suff}_i(K\mid \mathcal C_i,\Delta_i)=1\}.
\]
\(\mathsf{Ground}_i(d,K)\) checks whether \(K\) is supported by \(d\), either verbatim or as a mathematically faithful restatement. Grounding is a provenance predicate, not an independent verification that the source theorem is true. \(\mathsf{Suff}_i(K\mid \mathcal C_i,\Delta_i)\) checks whether \(K\), together with the proof context, justifies the transition; stronger statements receive credit only when their additional assumptions are satisfied. Thus correctness is membership in \(\mathcal V_i\), not equality to \(d_i^{\mathrm{cite}}\).

\subsection{Metrics}
\label{subsec:evaluation}

For an evaluation split \(\mathcal T\), the primary metric is
\[
\mathrm{ToolAcc}(\pi;\mathcal T)
:=
\frac{1}{|\mathcal T|}
\sum_{i\in\mathcal T}
\mathbf 1[(\hat d_i,\hat K_i)\in\mathcal V_i].
\]

We also report a Raw-track planning diagnostic, \(\mathrm{AnchorAcc}\). Let
\(
\widehat a^{\,\mathrm{raw}}_i(m)
:=
\pi_{\mathrm{plan}}
\big(
\mathcal C^i_{\mathrm{global}},
\mathcal C^i_{\mathrm{local}}(m)
\big)
\)
be the proof-intention description produced before retrieval, and let
\(\mathsf{AnchorMatch}_i(\widehat a,a_i)=1\) mean that \(\widehat a\) captures the same immediate proof role as the stored sanitized anchor \(a_i\), up to benign paraphrase. Then
\[
\mathrm{AnchorAcc}_{m}(\pi;\mathcal T)
:=
\frac{1}{|\mathcal T_a|}
\sum_{i\in\mathcal T_a}
\mathbf 1[
\mathsf{AnchorMatch}_i(\widehat a^{\,\mathrm{raw}}_i(m),a_i)=1
],
\qquad
\mathcal T_a:=\{i\in\mathcal T:a_i\neq\bot\}.
\]
The main paper reports \(\mathrm{AnchorAcc}:=\mathrm{AnchorAcc}_{5}\) with global context and the five-block local window. It measures proof-need abstraction before retrieval and is not a success criterion.

We report the grounding rate
\[
\mathrm{GroundRate}(\pi;\mathcal T)
:=
\frac{1}{|\mathcal T|}
\sum_{i\in\mathcal T}
\mathbf 1[\mathsf{Ground}_i(\hat d_i,\hat K_i)=1],
\]
which measures source-supported extraction without requiring proof-gap sufficiency.

Finally, to analyze citation-side retrieval without making it the target, let \(d\equiv_{\mathrm{src}}\mathsf{z}_i\) denote source identity between a retrieved candidate and the stored citation metadata. Then
\[
\mathrm{CiteRecall@}N
:=
\frac{1}{|\mathcal T|}
\sum_{i\in\mathcal T}
\mathbf 1\!
\left[
\exists d\in\mathcal R_{i,N}: d\equiv_{\mathrm{src}}\mathsf{z}_i
\right].
\]
CiteRecall@\(N\) asks whether the author-used construction witness appears in the retrieved set. It remains diagnostic only: the official correctness criterion is
\(
(\hat d_i,\hat K_i)\in\mathcal V_i.
\)
Implementation details for the retrieval contract, source-identity relation, judging rubric, and auxiliary diagnostics are given in Section~\ref{sec:experiment} and Appendix~\ref{app:evaluation_details}.
\section{Benchmark Construction}
\label{sec:dataset}

\ourbench{} instantiates the task in Section~\ref{sec:formulation}: each row stores a proof state \(\mathcal C_i\), an optional anchor \(a_i\), a reference witness \(K_i^\star\), and citation metadata \(d_i^{\mathrm{cite}}\); evaluation asks for any valid source--tool pair \((\hat d_i,\hat K_i)\in\mathcal V_i\). We summarize the construction, snapshot, and domain coverage here, and defer prompts, matching logic, curation rules, release details and anonymous release URL to Appendix~\ref{app:construction}.

\begin{figure*}[t]
  \centering
  \includegraphics[width=\textwidth]{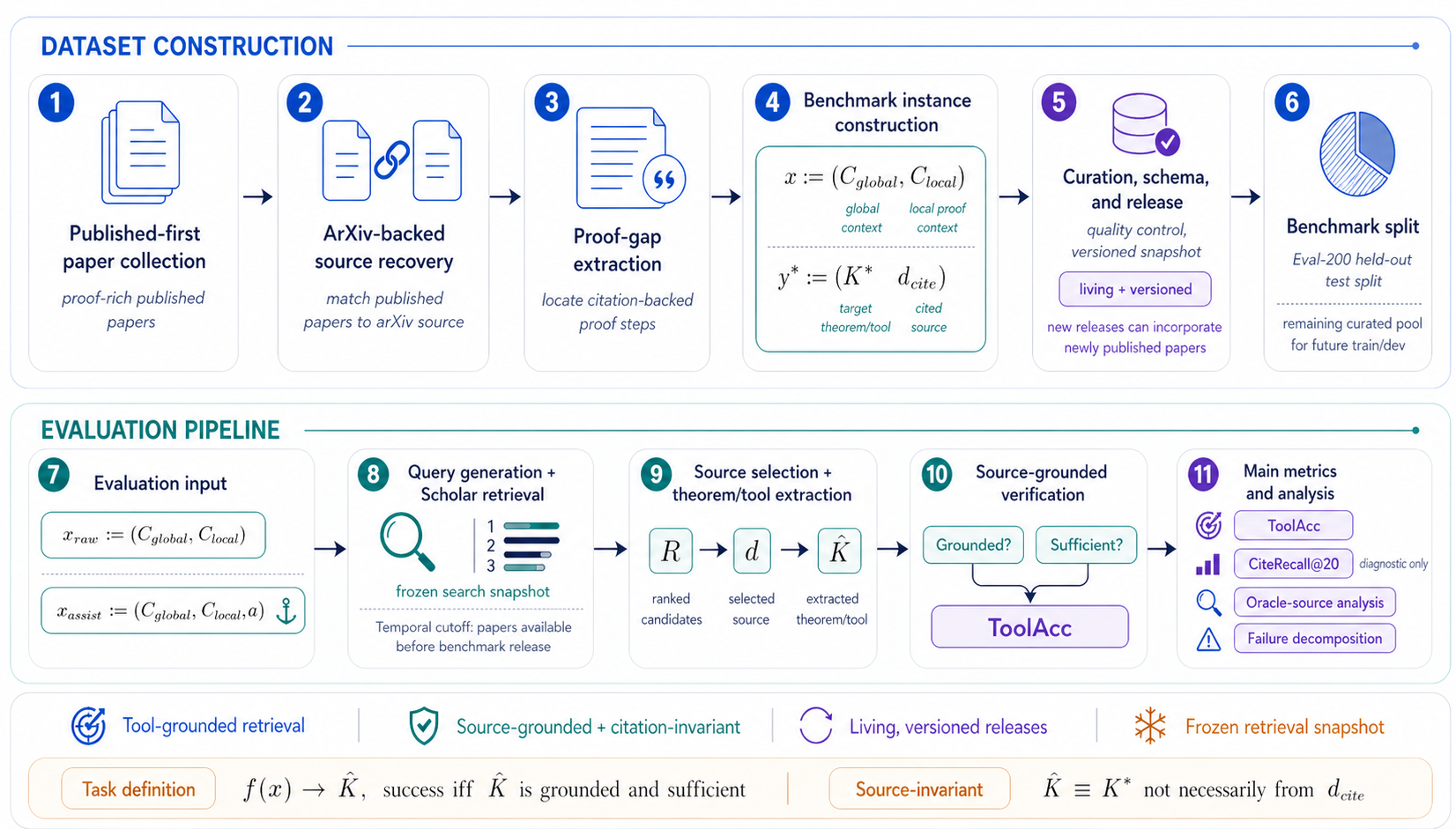}
  \caption{\textbf{Overview of \ourbench.}
A published-first, arXiv-backed pipeline mines candidate instrumental citations inside main-theorem proofs and converts them into proof-gap instances. Evaluation follows planning or anchor use, query generation, source selection, and theorem/tool extraction against a release-frozen retrieval artifact.}
  \label{fig:workflow}
\end{figure*}

\subsection{High-Level Pipeline}
\label{subsec:construction_strategy}

The pipeline is \emph{published-first, arXiv-backed}: we select proof-rich published papers, preserving their scholarly identity and citation structure, and use high-confidence arXiv twins with \LaTeX{} source when available to localize theorem statements, proof spans, and citation occurrences. This keeps the benchmark grounded in published mathematical writing while making extraction repeatable and refreshable.

For each accepted published--arXiv pair, a paper-level stage identifies the setup, target theorem, main-theorem proof span, and candidate instrumental citations. A citation-level stage converts retained citation occurrences into benchmark instances by extracting the pre-citation proof state \(\mathcal C_i\), a leakage-controlled anchor \(a_i\), a reference witness \(K_i^\star\), and auxiliary citation metadata \(d_i^{\mathrm{cite}}\). Ambiguous matches, low-confidence alignments, and suspicious witnesses are filtered during curation rather than forced into the release.

\subsection{Instance Format and Running Example}
\label{subsec:instance_format}

Each instance stores the formal objects used by the task: the agent input \(\mathsf{x}_i\), the reference witness \(K_i^\star\), and auxiliary citation metadata \(d_i^{\mathrm{cite}}\). The input consists of global context \(\mathcal C^i_{\mathrm{global}}\), a canonical local proof window \(\mathcal C^i_{\mathrm{local}}(5)\), and optionally the anchor \(a_i\) for the Assisted track. Retrieval-time artifacts such as \(Q_i\), \(\mathcal R_{i,N}\), \(\hat d_i\), and \(\hat K_i\) are produced by the evaluated agent, not stored as labels.

\paragraph{Running example.}
In one algebraic \(K\)-theory instance, \texttt{2110.11087v2\_gap\_1}, the global context defines \(K_2^G(R)=\ker(\mathrm{St}(\Phi,R)\to G(R))\) for Chevalley--Demazure groups and aims to prove \(K_2^G(R[t])\cong K_2^G(R)\). The local proof has handled axioms CFC and DP and leaves a type-\(A\) gluing step. The Assisted anchor asks for the polynomial gluing property of Steinberg groups; the reference witness \(K_i^\star\) states that \(\mathrm{St}(A_\ell,R[t])\) satisfies this property for regular \(R\supset k\), \(\ell\ge4\); and the author-cited witness is Tulenbaev's 1983 paper. A prediction succeeds only if it returns an admissible source supporting a sufficient gluing statement, not necessarily the citation key \texttt{Tu83}. Appendix~\ref{app:case_tulenbaev_gluing} gives the full normalized record.

\subsection{Current Snapshot and Domain Coverage}
\label{subsec:snapshot_domain_coverage}

The current release provides a snapshot of \ourbench{}, containing 860 proof-gap instances from 631 distinct papers after arXiv-version normalization. The dataset spans five domains: geometry/topology, analysis/PDE, probability/statistics/control, algebra/number theory, and combinatorics/discrete mathematics.
For evaluation, we construct a balanced Eval-200 test split with 40 instances per domain. Eval-200 is quality-controlled rather than random: it prioritizes journal-paper witnesses, nonempty
local context, anchors subjected to leakage controls, and complete reference witnesses; Appendix B.6
reports a separate construction audit of residual citation-role and anchor-sanitization risks.
\ourbench{} is designed as a versioned, continuously expanding benchmark. Future releases may extend the instance set and domain coverage by incorporating newly published
papers through the same automated pipeline. \emph{The pipeline is designed to be largely automated and low-friction, supporting incremental expansion with minimal additional curation effort.} An overview of the current dataset statistics is illustrated in Figure~\ref{fig:dataset_statistics}.

\section{Experiments}
\label{sec:experiment}

The experiments instantiate the source--tool contract in Section~\ref{sec:formulation}. We evaluate a fixed-budget research-assistance diagnostic: can a model map a proof state to a verifiable mathematical tool when all systems share the same information-access boundary? Models write the queries, but all retrieval is executed through the same release-frozen backend. This keeps retrieval inside the task while preventing model-native web-search, browser, file-search, or deep-research stacks from becoming hidden confounders. 

\subsection{Evaluation Conditions}
\label{subsec:exp_conditions}

\paragraph{Canonical end-to-end.}
The primary condition is the Raw track with global context and the stored five-block local window. For each instance, the model infers the proof need, emits search queries, observes the Top-20 metadata returned by the shared backend, selects at most one resolvable source to open in full text, and returns \((\hat d_i,\hat K_i)\). This is the main leaderboard condition because it tests the complete policy: planning, query generation, source selection, and theorem/tool extraction. In this condition, a completed run uses one Scholar query and at most one full-text source opening per
instance; uncertain, abstained, or non-evaluable judge outcomes are counted as negative.

\paragraph{Oracle-source diagnostic.}
To separate source acquisition from source use, we also run an oracle-source condition. When the author-cited source can be resolved and materialized under the release protocol, the model receives that source directly and must still localize, restate, and apply the relevant tool. \(\mathrm{OracleCoverage}\) is the fraction of Eval-200 for which this cited source is materializable, and oracle \(\mathrm{ToolAcc}\) is computed only on that subset. This diagnostic probes within-source reading and sufficiency checking; it is not a strict upper bound, because end-to-end systems may legitimately succeed through different admissible sources.

\paragraph{Auxiliary diagnostics.}
Appendix~\ref{app:additional_results} reports planning-window and assisted-query ablations, uncertainty intervals, domain stratification, source-level failure decomposition, and a single-annotator audit of judge labels. These diagnostics explain where failures enter the pipeline, but they do not replace the primary metric: whether the final source--tool pair belongs to \(\mathcal V_i\).

\subsection{Common Setting}
\label{subsec:exp_setup}

We evaluate on the official held-out Eval-200 split of the current release. The remaining curated instances are reserved as a non-test pool for future train/dev use and are not used for training, prompt tuning, or model selection in this paper. Splits are defined at the citing-paper level, so multiple gaps from the same paper cannot cross partitions.

\paragraph{Models.}
The paper-facing results report seven completed model runs: GPT-5.2, Gemini 3.1 Pro, Claude Opus 4.5, DeepSeek V3.2, Qwen3-235B Thinking, Kimi K2 Thinking, and Grok 4. Each reported run uses the same benchmark interface, retrieval budget, source-materialization policy, and judge rubric.

\paragraph{Shared retrieval artifact.}
All models use the same release-frozen retrieval artifact, instantiated with Google Scholar. We use Google Scholar as a broad scholarly discovery interface because the needed tool may appear across heterogeneous records---articles, proceedings papers, preprints, books, surveys, or bibliographic entries---rather than in an arXiv-only or publisher-specific index. For each proof gap, model-authored queries are executed once under fixed externally controlled Google Scholar settings, with no lower-bound year and an upper-bound year given by the original citing paper's publication year. The evaluator caches the returned Top-20 metadata, rank order, query timestamp, and source-materialization outcomes when attempted. The benchmark therefore fixes two things: a per-instance bibliographic cutoff to avoid later-literature leakage, and a frozen query-result artifact for reproducibility. Since Google Scholar's underlying coverage and ranking are inherited rather than re-indexed by us, \(\mathrm{CiteRecall@20}\) is a query--artifact diagnostic, while \(\mathrm{ToolAcc}\) remains the primary source--tool metric.

\paragraph{Judge.}
Outputs are scored by a held-out non-roster judge model fixed across runs. The judge applies the two predicates in Section~\ref{sec:formulation}: grounding in the selected source and sufficiency for the proof transition. Appendix~\ref{app:evaluation_details} gives the prompt rubrics, and Appendix~\ref{app:additional_results} reports uncertainty diagnostics together with a single-annotator human audit. The audit is judge-vs-human-confirmed validation, not an inter-annotator agreement study.

\paragraph{Metric.}
The main paper reports the primary end-to-end metric, \(\mathrm{ToolAcc}\), together with
three diagnostics: \(\mathrm{AnchorAcc}\), \(\mathrm{CiteRecall@20}\), and
\(\mathrm{GroundRate}\) defined in Section~\ref{subsec:evaluation}. 

The oracle table additionally reports \(\mathrm{OracleCoverage}\), \(\mathrm{Oracle\ ToolAcc}\), the oracle
gap, and alternative-source diagnostics. \(\mathrm{OracleCoverage}\) is the fraction of Eval-200 instances
for which the author-cited source could be resolved and materialized in a completed oracle run; it is
a release-protocol materialization property, not a model score. \(\mathrm{Oracle\ ToolAcc}\) is computed
on the materialized subset for each completed run and should be read as an upper-bound-style
diagnostic rather than a perfectly controlled same-subset counterfactual. We read the metrics as a
staged diagnosis: \(\mathrm{AnchorAcc}\) probes proof-need abstraction, \(\mathrm{CiteRecall@20}\) probes
whether the stored witness appears in the retrieved pool, \(\mathrm{GroundRate}\) probes source-supported
extraction, \(\mathrm{ToolAcc}\) probes mathematical sufficiency, oracle evaluation probes within-source
reading, and alternative-source metrics test the citation-invariant design.

\section{Results and Analysis}
\label{sec:analysis}

Under the fixed-budget shared-backend setting, current frontier LLMs are not reliable at tool-grounded mathematical retrieval.  The strongest fixed-judge score is only \(7.0\%\) \(\mathrm{ToolAcc}\), and all
other fixed-judge scores fall between \(1.0\%\) and \(3.5\%\).  We therefore treat the leaderboard as a diagnostic rather than a ranking endpoint: the main signal is the failure structure.  Table~\ref{tab:main_core_e2e} already shows this structure.  Kimi has the strongest planning signal, Grok has the highest grounding rate, but Claude has the best end-to-end \(\mathrm{ToolAcc}\).  Intermediate progress is not a monotone proxy for source--tool success.

\begin{table*}[t]
\centering
\small
\renewcommand{\arraystretch}{1.10}
\setlength{\tabcolsep}{4.8pt}
\caption{\textbf{Core end-to-end results on Eval-200.} Values are percentages with successful counts in parentheses; all denominators are 200. Models are sorted by the primary metric, \(\mathrm{ToolAcc}\). Bold marks the best value and underline marks the second best.}
\label{tab:main_core_e2e}
\begin{tabular*}{\textwidth}{@{\extracolsep{\fill}}lcccc@{}}
\toprule
& \multicolumn{1}{c}{Planning} & \multicolumn{1}{c}{Citation-side} & \multicolumn{1}{c}{Grounding} & \multicolumn{1}{c}{End-to-end} \\
\cmidrule(lr){2-2}\cmidrule(lr){3-3}\cmidrule(lr){4-4}\cmidrule(l){5-5}
\textbf{Model} & \(\mathrm{AnchorAcc}\uparrow\) & \(\mathrm{CiteRecall@20}\uparrow\) & \(\mathrm{GroundRate}\uparrow\) & \(\mathrm{ToolAcc}\uparrow\) \\
\midrule
Claude Opus 4.5     & 24.0\% (48)                & \textbf{10.5\%} (21)      & 24.0\% (48)             & \textbf{7.0\%} (14) \\
Grok 4              & 26.5\% (53)                & 4.5\% (9)                 & \textbf{41.5\%} (83)    & \underline{3.5\%} (7) \\
Kimi K2 Thinking    & \textbf{43.5\%} (87)      & 6.0\% (12)                & 24.0\% (48)             & \underline{3.5\%} (7) \\
GPT-5.2             & 5.5\% (11)                 & \underline{8.5\%} (17)    & 25.0\% (50)             & 3.0\% (6) \\
DeepSeek V3.2       & 8.0\% (16)                 & 5.5\% (11)                & \underline{29.0\%} (58) & 2.5\% (5) \\
Gemini 3.1 Pro      & 2.0\% (4)                  & 5.5\% (11)                & 12.0\% (24)             & 2.0\% (4) \\
Qwen3-235B Thinking & \underline{40.5\%} (81)   & 6.0\% (12)                & 17.0\% (34)             & 1.0\% (2) \\
\bottomrule
\end{tabular*}
\begin{minipage}{0.98\textwidth}
\footnotesize
\end{minipage}
\end{table*}

\paragraph{Grounding is much easier than applicability.}
Models often retrieve or restate true mathematics without making it usable.  Grok grounds statements on \(83/200\) instances but succeeds on only \(7/200\); DeepSeek grounds \(58/200\) but succeeds on \(5/200\); GPT-5.2 grounds \(50/200\) but succeeds on \(6/200\).  These are often not hallucinations, but true, sourced, and thematically nearby statements that do not discharge the proof obligation.  The missing capability is assumption accounting: checking objects, domains, hypotheses, parameter ranges, quantifiers, and conclusion strength against the local proof state.

\paragraph{Proof intent must survive a keyword-oriented search interface.}
Understanding the proof need is not equivalent to retrieving the right document.  The shared scholarly backend is closer to keyword and bibliographic matching than to semantic theorem retrieval.  A model must therefore express the proof obligation in terms the index can surface.  The Steinberg gluing case shows that even natural phrases such as ``Steinberg group polynomial gluing property'' may fail to surface a usable source; the Muskat case shows a correct compactness need collapsing into generic ``Aubin--Lions'' search terms.  Appendix~\ref{app:case_muskat_query_collapse} gives finer lexicalization probes.

\begin{table*}[t]
\centering
\scriptsize
\renewcommand{\arraystretch}{1.10}
\setlength{\tabcolsep}{3.5pt}
\caption{\textbf{Oracle-source and citation-invariant diagnostics.} Oracle-source evaluation provides the author-cited source when it can be materialized, isolating within-source localization and extraction. Alternative-source metrics count valid successes through non-cited sources. Oracle values are upper-bound-style diagnostics from the completed oracle runs, not perfectly controlled same-subset reruns.}
\label{tab:main_oracle}
\begin{tabular*}{\textwidth}{@{\extracolsep{\fill}}lcccccc@{}}
\toprule
& \multicolumn{1}{c}{End-to-end} & \multicolumn{3}{c}{Author-source oracle} & \multicolumn{2}{c}{Citation-invariant success} \\
\cmidrule(lr){2-2}\cmidrule(lr){3-5}\cmidrule(l){6-7}
\textbf{Model} &
\(\mathrm{ToolAcc}\) &
\shortstack[c]{Coverage} &
\shortstack[c]{Oracle\\\(\mathrm{ToolAcc}\)} &
\shortstack[c]{\(\Delta\)\\(pp)} &
\shortstack[c]{Alt. source\\\(\mathrm{ToolAcc}\)} &
\shortstack[c]{Alt. share\\among successes} \\
\midrule
Claude Opus 4.5  & \textbf{7.0\%} (14/200)  & 35.5\% (71/200) & \textbf{12.7\%} (9/71)  & +5.7 & \textbf{6.5\%} (13/200)    & 92.9\% (13/14) \\
Grok 4           & \underline{3.5\%} (7/200) & 37.0\% (74/200) & 2.7\% (2/74)            & -0.8 & \underline{3.5\%} (7/200) & 100.0\% (7/7) \\
Kimi K2 Thinking & \underline{3.5\%} (7/200) & 37.0\% (74/200) & 1.4\% (1/74)            & -2.1 & \underline{3.5\%} (7/200) & 100.0\% (7/7) \\
GPT-5.2          & 3.0\% (6/200)              & 35.5\% (71/200) & 7.0\% (5/71)            & +4.0 & 2.5\% (5/200)              & 83.3\% (5/6) \\
DeepSeek V3.2    & 2.5\% (5/200)              & 37.0\% (74/200) & 2.7\% (2/74)            & +0.2 & 2.5\% (5/200)              & 100.0\% (5/5) \\
Gemini 3.1 Pro   & 2.0\% (4/200)              & 33.5\% (67/200) & \underline{7.5\%} (5/67) & +5.5 & 1.5\% (3/200)              & 75.0\% (3/4) \\
Qwen3-235B Thinking & 1.0\% (2/200)           & 37.0\% (74/200) & 2.7\% (2/74)            & +1.7 & 1.0\% (2/200)              & 100.0\% (2/2) \\
\bottomrule
\end{tabular*}
\begin{minipage}{0.98\textwidth}
\footnotesize
\emph{Remark.} Coverage is a release-protocol materialization property, not a model score. Main values are fixed-judge benchmark scores. Appendix~\ref{subsec:appendix_judge_failure} reports a single-annotator audit sensitivity analysis identifying three alternative-source false positives; retroactive correction only lowers the affected \(\mathrm{ToolAcc}\) and alternative-source counts.
\end{minipage}
\end{table*}

\paragraph{Search helps, but source use remains hard.}
Table~\ref{tab:main_oracle} removes open-world source acquisition for materialized author sources, yet the best oracle score is only $12.7\%$ and several models remain below $3\%$.  Thus retrieval is not the only bottleneck: after a paper is found, the model must still locate the theorem boundary and verify that its assumptions and strength fit the proof transition.

\paragraph{Citation-invariant scoring is mathematically necessary.}
Most successful predictions do not recover the stored citation: Claude succeeds through alternative sources in \(13/14\) judge-scored cases, GPT-5.2 in \(5/6\), and Kimi, Grok, DeepSeek, and Qwen in all of their successful cases.  A citation-locked metric would mark many source-grounded tools wrong, yet alternative-source \(\mathrm{ToolAcc}\) is still at most \(6.5\%\).  A single-annotator audit in Appendix~\ref{subsec:appendix_judge_failure} validates \(39/42\) alternative-source successes and identifies three fixed-judge false positives; applying those corrections would only lower the reported scores, so the failure conclusion is unchanged.

\paragraph{Synthesis.}
The dominant failures are two interfaces: proof obligations must become searchable lexical evidence, and sourced theorems must become assumption-checked proof tools.  This motivates theorem-aware retrieval backends that index statement boundaries, assumptions, and conclusion structure under the same frozen-artifact contract.

\section{Conclusion}
\label{sec:conclusion}

We introduced \ourbench, a benchmark for source-grounded mathematical tool use from
research-proof context. Under the fixed-budget shared-backend setting, the strongest
fixed-judge result on Eval-200 is only \(7.0\%\) \(\mathrm{ToolAcc}\), even though
several models ground many more statements in real sources. This reveals a central gap
in current LLM-based research assistance: models can often find mathematics, but they
cannot reliably make it usable. The missing capability is the disciplined composition of
retrieval, provenance, theorem localization, and assumption checking into a proof-valid
source--tool pair. \ourbench\ turns this capability into a reproducible evaluation target
and points toward theorem-aware retrieval systems that index statements, hypotheses, and
conclusion structure rather than paper-level relevance alone.

\bibliographystyle{plainnat}
\bibliography{sample-base}


\appendix
\section{Additional Discussion: What the Benchmark Is Designed to Reveal}
\label{sec:discussion}

This section discusses the methodological implications of \ourbench independently of any specific model ranking. The goal is to clarify what kinds of capability gaps the benchmark is intended to expose, and why the current evaluation is organized around source-grounded success rather than citation recovery alone.

\subsection{Author Citations Provide Supervision, Not the Definition of Success}

Instrumental citations are a strong construction signal because they tell us that a human author relied on an external result at a specific proof transition. But the cited source is often not the only valid way to justify the step. The same theorem may appear in multiple papers, books, surveys, or lecture notes, and a stronger statement from a different source may also be perfectly usable.

This is why \ourbench is citation-invariant but source-grounded. A citation-locked benchmark would underestimate genuine literature competence by penalizing alternative valid sources. By contrast, our official success criterion requires only that the returned statement be grounded in the selected source and sufficient for the gap.

\subsection{Why We Separate Planning, Retrieval, and Within-Source Extraction}

Mathematical tool use is a pipeline, not a single decision. A system may fail because it cannot infer the right kind of next step from proof context, because it cannot turn that need into a useful search query, because it cannot recognize the right source from metadata, or because it cannot localize the correct statement inside the selected document.

The benchmark therefore fixes separate views of the problem. Raw-track ablations ask whether the hierarchical context supports the intended proof-level planning signal. End-to-end evaluation measures actual open-world retrieval and grounded extraction. Oracle-source evaluation then removes open-world retrieval and isolates within-source localization and sufficiency. This decomposition is designed to make future method improvements more attributable.

\subsection{Why Source-Level Failure Decomposition Matters}

A single low end-to-end score is hard to interpret.  It does not tell us whether the final output lacked any source support, whether it was grounded but mathematically insufficient, or whether the source--tool pair succeeded.  Because \ourbench{} is citation-invariant, the paper-facing failure decomposition is source-level rather than citation-conditioned: it partitions outputs by \textsc{Ground}, \textsc{Suff}, and \textsc{ToolAcc}, not by whether the author-cited source appears in Top-20.

This distinction is especially important in mathematics. A model may retrieve a thematically related source and extract a true theorem, yet the theorem may have incompatible assumptions, a weaker conclusion, or the wrong objects.  Retrieval-conditioned views based on \(\mathrm{CiteRecall@20}\) remain useful for debugging, but they are not the official failure definition.

\subsection{Deployment Boundary}

Even a strong score on \ourbench would not make a system an autonomous proof assistant in the formal sense. The benchmark evaluates source-grounded tool use under natural-language proof context, not mechanically checked proof completion. In practical workflows, the most immediate use case is therefore a research copilot that helps surface candidate tool families, likely sources, and plausible theorem statements for human verification.

At the same time, \ourbench offers a clearer target for future systems that combine retrieval, verification, and formalization. Because the task explicitly requires source grounding and applicability under context, it can serve as a bridge between open-world mathematical search and downstream formal proof checking.

\section{Dataset Construction Protocol}
\label{app:construction}

This appendix specifies the release protocol behind \ourbench.  The goal is to make the benchmark auditable along three axes: (i) how proof-gap instances are mined from published mathematical writing, (ii) how the release and evaluation split are frozen, and (iii) how model outputs are rendered, retrieved, judged, and aggregated.  We use the notation of Section~\ref{sec:formulation}; implementation-specific choices are described here rather than in the task definition.

\begin{figure}[t]
  \centering
  \includegraphics[width=\linewidth]{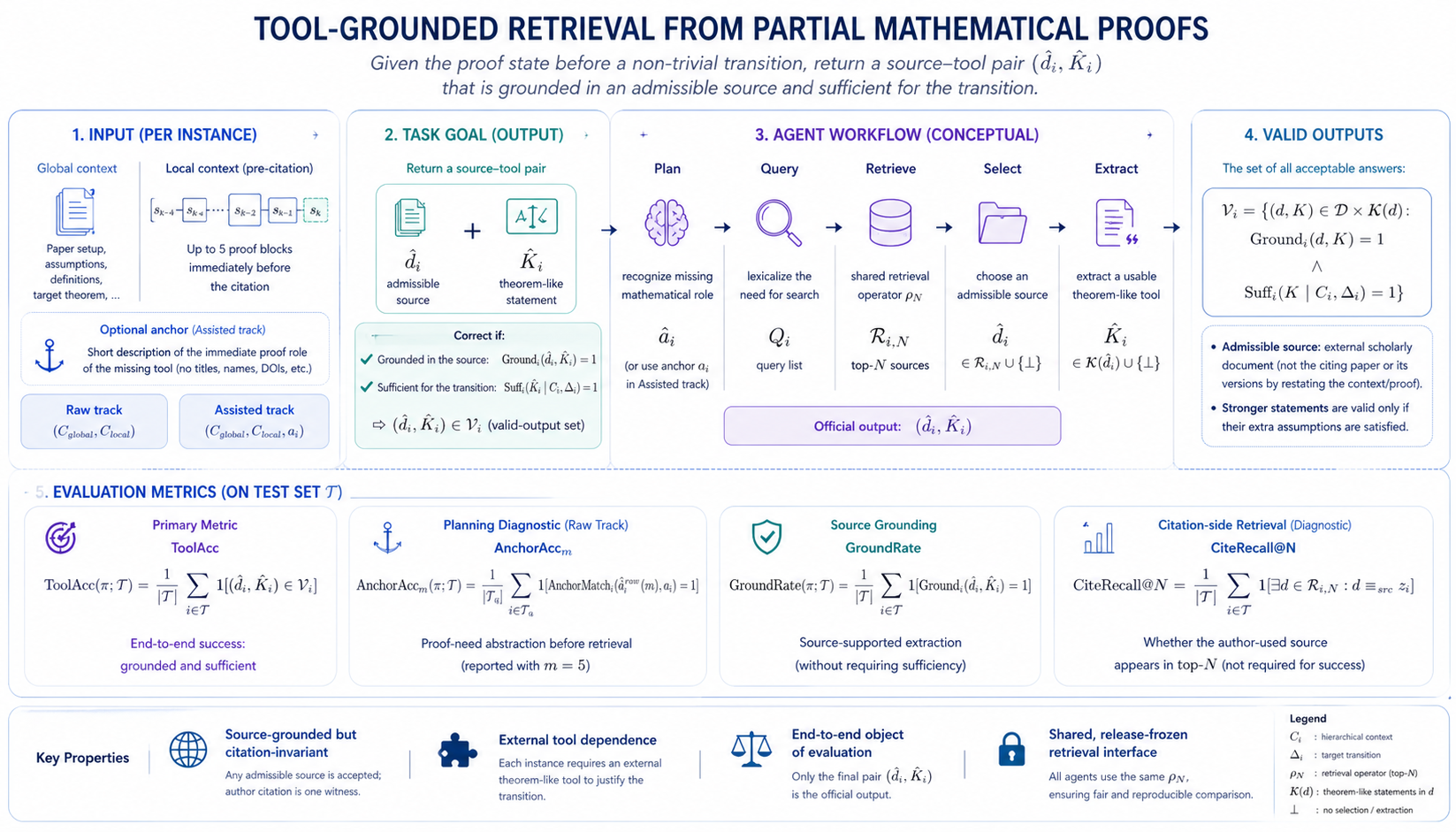}
  \caption{
  \textbf{Overview of tool-grounded retrieval.}
  Given hierarchical proof context (global + local) and an optional anchor, the agent produces a source--tool pair via planning, query formulation, retrieval, source selection, and statement extraction.
  A prediction is correct if the extracted statement is grounded in the selected source and sufficient for the target proof transition.
  The figure is illustrative; formal definitions are given in Section~\ref{sec:formulation}.
  }
  \label{fig:task_overview}
\end{figure}

\paragraph{Illustrative overview.}
For readers who prefer a high-level, visual summary, we provide an illustrative diagram of the task pipeline in Figure~\ref{fig:task_overview}. The figure depicts the end-to-end workflow—from proof context and (optional) anchor, through query formulation and retrieval, to source selection and tool extraction—and highlights the validity criterion based on grounding and proof-gap sufficiency.
This diagram is intended for intuition only and does not replace the formal task definition in Section~\ref{sec:formulation}, which specifies the precise inputs, outputs, and evaluation criteria.

\newtcolorbox{promptbox}[1]{%
  breakable,
  colback=blue!2,
  colframe=blue!45!black,
  boxrule=0.35pt,
  arc=1.2mm,
  left=0.9mm,
  right=0.9mm,
  top=0.7mm,
  bottom=0.7mm,
  fonttitle=\bfseries\footnotesize,
  before upper={\raggedright\small},
  title={#1},
  before skip=0.45em,
  after skip=0.55em
}

\subsection{Published-First Manifest and arXiv Pairing}
\label{subsec:appendix_manifest}

\ourbench\ is constructed under a \emph{published-first, arXiv-backed} policy.  Candidate citing papers are selected from the published literature, so each instance inherits a real venue, citation record, and scholarly identity.  When a high-confidence arXiv counterpart with \LaTeX{} source is available, we use that source for extraction, alignment, and proof-local parsing.  This separation lets the benchmark retain a published-literature identity while avoiding brittle OCR over publisher PDFs.

The manifest records two linked identities for every accepted paper.  The \emph{published identity} stores title, authors, venue, year, publication date, DOI, and related bibliographic metadata.  The \emph{extraction identity} stores the arXiv identifier, source URL, matched title, match type, and match confidence.  Matching proceeds through a conservative cascade: DOI-exact agreement, then high-confidence title--author--year agreement, then manual verification.  Ambiguous pairs are sent to a review queue rather than forced into the release.

\begin{table*}[t]
\centering
\small
\caption{\textbf{Construction protocol overview.}  The pipeline separates paper identity, proof-local mining, instance formation, and release curation.}
\label{tab:appendix_construction_protocol}
\renewcommand{\arraystretch}{1.12}
\setlength{\tabcolsep}{4.0pt}
\resizebox{\textwidth}{!}{%
\begin{tabular}{@{}p{0.18\textwidth}p{0.28\textwidth}p{0.30\textwidth}p{0.16\textwidth}@{}}
\toprule
\textbf{Component} & \textbf{Input} & \textbf{Operation} & \textbf{Release artifact} \\
\midrule
Manifest & Published mathematical papers & Match to arXiv source when available; preserve both identities & Published--arXiv pair \\
Paper scouting & Matched \LaTeX{} source and bibliography & Locate setup, target theorem proof, and instrumental proof citations & Proof-local citation candidates \\
Citation extraction & Citation-local proof slice & Build pre-citation context, anchor, reference witness, and citation metadata & Canonical row \\
Curation & Candidate rows & Deduplicate, repair, filter, and split at citing-paper level & Frozen release and Eval-200 \\
\bottomrule
\end{tabular}}
\end{table*}

\subsection{Two-Stage Mining Pipeline}
\label{subsec:appendix_two_stage_mining}

\paragraph{Stage I: paper-level proof scouting.}
The first stage operates at the paper level.  Given a matched arXiv source, a compact bibliography map, and a target-theorem hint, the scout extracts three objects: a global setup, the proof span of the target theorem, and proof-internal citation occurrences that appear to function as external mathematical tools.  This stage deliberately excludes historical remarks, background citations, definitions, and citations outside the target proof.

\paragraph{Stage II: citation-level instance extraction.}
The second stage processes one retained citation occurrence at a time.  It aligns the citation inside a proof-local slice, extracts the ordered pre-citation blocks, and constructs the triple
\[
(\mathsf{x}_i,\mathsf{y}_i,\mathsf{z}_i)
=
\big((\mathcal C^i_{\mathrm{global}},\mathcal C^i_{\mathrm{local}},a_i),
K_i^\star,
d_i^{\mathrm{cite}}\big)
\]
used in the main text.  If the citation cannot be localized reliably, the candidate is dropped.  We do not backfill missing local context from unrelated parts of the proof.

\subsection{Anchor Resolution and Leakage Controls}
\label{subsec:appendix_anchor_controls}

The anchor hint \(a_i\) is a planning cue, not an answer.  It is produced by combining an LLM-generated proof-intention description, citation-local heuristics, implication-style rewrites from the locator snippet, and local-context heuristics.  The final anchor is designed to be leakage-controlled: it may describe the mathematical role of the missing tool, but it should not include theorem titles, author names, citation keys, source titles, DOI strings, arXiv identifiers, or other bibliographic shortcuts.  Manual repairs are applied only to remove leakage, repair malformed text, or make the intended proof move grammatically clear.

\subsection{Construction Prompt Templates}
\label{subsec:appendix_construction_prompts}

The construction prompts below are written as release-facing prompt contracts.  The actual released code may call fallback variants depending on parse failures or weak tool statements, but all variants preserve the same principle: recover proof-local external tool use without inventing mathematical content.

\begin{promptbox}{C1. Stage-1 structure and citation scouting}
\small
\textbf{Role.} Extract one benchmark instance from one mathematical paper. Return JSON only.

\textbf{Input payload.} A compact bibliography map and a compact paper view constructed from the matched arXiv source.

\textbf{Required fields.}
\begin{itemize}
\item \texttt{global\_context.setup}: concise setup needed for the paper's main theorem.
\item \texttt{global\_context.target\_theorem}: the main theorem only; return an empty string if uncertain.
\item \texttt{proof\_span.start\_snippet} and \texttt{proof\_span.end\_snippet}: verbatim 20--40 word snippets delimiting the proof of the target theorem when recoverable.
\item \texttt{proof\_citations}: at most six citations used as external mathematical tools inside that proof.  Each item contains a citation key, locator snippet, and short usage reason.
\end{itemize}

\textbf{Rules.} Use only citation keys from the bibliography map. Exclude citations used only for background, setup, history, definitions, or non-proof discussion. Prefer empty fields over guessing. Keep the setup and target theorem compact.
\end{promptbox}

\begin{promptbox}{C1-fallback. Stage-1 recall and repair prompts}
\small
\textbf{Compact retry.} When a reasoning model truncates or fails to emit valid JSON, return one compact JSON object with the same keys as C1, no reasoning, no markdown, and at most four proof citations.

\textbf{Targeted recall.} When the proof span or proof citations are missing, act as a Mathematical Proof Citation Scout. Given a target-theorem hint, optional setup, bibliography map, and compact paper view, recover the proof span and recall plausible external-tool citations inside that proof. Favor recall over precision, but discard background, historical, definitional, or non-proof citations.

\textbf{Proof-local recall.} When the proof span has already been identified, read only the provided proof text and recover citations used as external theorems, lemmas, propositions, criteria, estimates, inequalities, or structural facts. Return JSON with a \texttt{proof\_citations} list.

\textbf{Structured repair.} When a prompt output is malformed, repair syntax into the requested JSON schema while preserving content when possible. Do not invent new mathematical content beyond minimal syntax repair.
\end{promptbox}

\begin{promptbox}{C2. Stage-2 citation-local witness extraction}
\small
\textbf{Role.} Mathematical Logic Expert.

\textbf{Input payload.} Target theorem, target citation key, pre-citation local proof blocks, citation-focus snippet, short tool-usage reason, and bibliography entry.

\textbf{Task.} Recover only the grounded external tool for the given citation occurrence. Focus on the proof step immediately around the citation. Return the minimal theorem-like statement, estimate, implication, or criterion needed at this step.

\textbf{Rules.}
\begin{itemize}
\item Do not rewrite the local context blocks.
\item Do not output a theorem title alone.
\item Prefer an explicit implication, estimate, criterion, or theorem-like statement over vague prose.
\item Reject discourse wrappers such as ``Recall that,'' ``In addition,'' case-analysis narration, future-work text, examples, and remarks.
\item Set \texttt{restated\_in\_citing\_paper=true} only when the citing text explicitly states the mathematical implication or estimate itself.
\item The returned statement must not include theorem labels or source-reporting wrappers such as ``Proposition 1.5 states that.''
\end{itemize}

\textbf{Output.} JSON with \texttt{reference\_tool\_latex}, \texttt{reference\_tool\_type}, and \texttt{restated\_in\_citing\_paper}.
\end{promptbox}

\begin{promptbox}{C2-fallback. Focused tool extraction and implication rewrite}
\small
\textbf{Trigger.} Used when the default Stage-2 output is empty, generic, wrapped in narrative text, or not theorem-like enough for a benchmark row.

\textbf{Primary retry.} Focus only on recovering the external tool statement used by the specific citation. If the paper only reveals the consequence used at this step, return that consequence as a faithful theorem-like paraphrase. Do not return an empty string unless no recoverable theorem-like content is provided.

\textbf{Implication pass.} If the citation sentence says that an object belongs to one class and therefore belongs to another class by a cited result, rewrite the operative content as a theorem-like implication of the form ``If ... then ...''. A short usable implication is preferred over an empty answer.

\textbf{Batch extraction.} When enabled, the same fields are extracted for multiple citation candidates jointly; each item still returns a citation key, tool statement, tool type, and restatement flag.
\end{promptbox}

\subsection{Curation, Schema, Release, and Split Policy}
\label{subsec:appendix_release_policy}

After mining, the curation step deduplicates candidate rows, repairs broken core fields, removes low-confidence alignments, and verifies that all required task fields are present.  The canonical schema follows the \(\mathsf{x}/\mathsf{y}/\mathsf{z}\) split in Section~\ref{sec:formulation}: \(\mathsf{x}\) stores the proof context and optional anchor, \(\mathsf{y}\) stores the reference theorem-like witness, and \(\mathsf{z}\) stores auxiliary citation metadata.

The local proof context stores at most the last five pre-citation proof blocks in order.  Local-window analyses are obtained by slicing this stored list from the end with \(m\in\{1,3,5\}\).  If fewer than \(m\) blocks are available, the evaluator uses the shorter available context; if no pre-citation block exists, the local context remains empty rather than fabricated.

Each paper-facing comparison is tied to a frozen release: a fixed instance set, an official held-out evaluation split, retrieval provider and externally controlled settings, query log, and cached candidate metadata.  The balanced Eval-200 subset is the official held-out evaluation split used in the main experiments.  The remaining curated rows form a non-test pool for future train/dev use.  The current paper does not train, tune, or select prompts on Eval-200.  Future partitions should be made at the citing-paper level rather than the proof-gap level, because a single citing paper may contribute multiple related gaps.

\paragraph{Eval-200 selection.}
The held-out evaluation split is quality-filtered rather than randomly sampled.  The selection script requires a journal-paper cited source, nonempty locator, nonempty leakage-controlled anchor, nonempty complete reference witness, and nonempty local context.  It then ranks candidates by a quality score, enforces a per-paper cap, and fills a balanced number of instances in each benchmark domain.

\paragraph{Leaderboard and release policy.}
A public comparison should identify the full frozen tuple: dataset release, evaluation split, retrieval provider and externally controlled settings,
retrieval/query-cache artifact, source-materialization log, scoring artifact, and evaluation date.  A future model may generate queries that are absent from an earlier cache; in that case, the benchmark records a new query-result artifact under the same retrieval provider and externally controlled settings rather than silently mixing live search with cached results.  Results from different dataset releases or retrieval artifacts should not be treated as entries on the same leaderboard unless older systems are re-evaluated under the same artifact.

\begin{table*}[t]
\centering
\small
\caption{\textbf{Release and leaderboard contract.}  This table records the artifact components that make a living benchmark comparable across submissions.}
\label{tab:appendix_release_contract}
\renewcommand{\arraystretch}{1.10}
\setlength{\tabcolsep}{4pt}
\resizebox{\textwidth}{!}{%
\begin{tabular}{@{}p{0.22\textwidth}p{0.38\textwidth}p{0.32\textwidth}@{}}
\toprule
\textbf{Artifact} & \textbf{Frozen contents} & \textbf{Comparison rule} \\
\midrule
Dataset release & Instance set, schema, split, domain labels, construction manifest & Results are tied to a named release. \\
Retrieval artifact & Retrieval provider and externally controlled settings, evaluated queries,
Top-20 candidate metadata, timestamps,
aggregation rules & New query logs require a new frozen artifact or re-evaluation. \\
Source materialization log & Resolved source identifiers, full-text access status, parsing status & Oracle coverage is interpreted as materialization coverage, not model accuracy. \\
Scoring artifact & Judge version, rubric prompts, source matcher, aggregation script & Scores should be recomputed if the judge or matcher changes. \\
\bottomrule
\end{tabular}}
\end{table*}

\paragraph{Anonymous release access.}
An anonymous reviewer-accessible release is available at
\url{https://github.com/Anonymous1-bot-bit/Re-2Math}.
The release contains the curated instances, Eval-200 split, non-test pool,
schema, retrieval/query-cache artifact, source-materialization log, scoring
artifact, prompt contracts, aggregation scripts, and Croissant metadata. The
release will be made public and de-anonymized by the camera-ready deadline if
accepted.

\subsection{Construction Quality Audit}
\label{subsec:construction_quality_audit}

Because the construction pipeline is automated and LLM-assisted, we audited a fixed subset of 120 independent retained instances.  The audit checks whether each row is genuinely proof-local, externally tool-dependent, and leakage-controlled.  This is a single-annotator construction audit rather than an inter-annotator reliability study.  Table~\ref{tab:construction_quality_audit} summarizes the results.

\begin{table*}[t]
\centering
\small
\caption{\textbf{Single-annotator construction audit.}   The audit covers 120 independent retained instances
and is used to characterize construction risk, not to rescore Eval-200. It validates local alignment,
reference-witness sufficiency, metadata, and external-source dependence, while identifying
citation-role instrumentality and anchor sanitization as the main residual construction risks.}
\label{tab:construction_quality_audit}
\renewcommand{\arraystretch}{1.10}
\setlength{\tabcolsep}{5pt}
\resizebox{\textwidth}{!}{%
\begin{tabular}{@{}p{0.28\textwidth}p{0.38\textwidth}p{0.16\textwidth}p{0.16\textwidth}@{}}
\toprule
\textbf{Audit item} & \textbf{Pass condition} & \textbf{Positive / audited} & \textbf{Rate} \\
\midrule
Instrumental citation & Citation supplies a non-trivial external tool for the local proof transition & 110 / 120 & 91.7\% \\
Local-context alignment & Stored pre-citation blocks are the exposed proof state for the target citation & 120 / 120 & 100.0\% \\
Reference witness sufficiency & Curated \(K_i^\star\) is sufficient for the gap under the stored context & 120 / 120 & 100.0\% \\
Anchor leakage control & Anchor contains no theorem title, author, citation key, DOI, arXiv ID, or source title & 98 / 120 & 81.7\% \\
Citation metadata correctness & Stored \(d_i^{\mathrm{cite}}\) matches the source used by the citing paper & 116 / 120 & 96.7\% \\
External-source constraint & Instance requires an external tool rather than merely restating the citing proof & 120 / 120 & 100.0\% \\
\bottomrule
\end{tabular}}
\end{table*}

The two weakest dimensions are citation-role instrumentality and anchor sanitization. We interpret
these flags conservatively. Instrumentality is a construction-signal audit: a failure means that the
original citation is not a clean witness for the local proof transition, not that the benchmark row has
no valid source--tool solution. In the same audit, all 120 rows passed reference-witness sufficiency
and the external-source constraint, so the scored task remains the final source--tool predicate rather
than exact recovery of the author citation.

There were 10 instrumentality flags. In a triaged subset of five flagged cases, the issues consisted of
background/historical citation roles (3) and comparison/related-work citation roles (2); no pure
definition-citation or locally self-contained proof-step case was observed in that triage. The remaining
flagged cases are therefore treated conservatively as citation-role ambiguity. Under the frozen-release
contract, post-hoc removal of rows after evaluation would define a new dataset release and require
corresponding retrieval and scoring artifacts. We therefore report these flags as residual construction
noise in the author-citation witness and use them to tighten future citation filters, rather than changing
the fixed Eval-200 denominator post hoc.

Anchor-sanitization failures are handled separately because anchors are not exposed in the Raw
end-to-end leaderboard. They can affect Assisted-track query diagnostics and may make stored anchors
imperfect as planning targets; accordingly, assisted-query results are reported only as diagnostics and
not as official task scores. The validity predicates Ground and Suff are evaluated on the model's
selected source--tool pair and do not use anchor text as evidence.

\subsection{Snapshot Statistics}
\label{subsec:appendix_dataset_stats}

\begin{figure}[t]
    \centering
    \includegraphics[width=\linewidth]{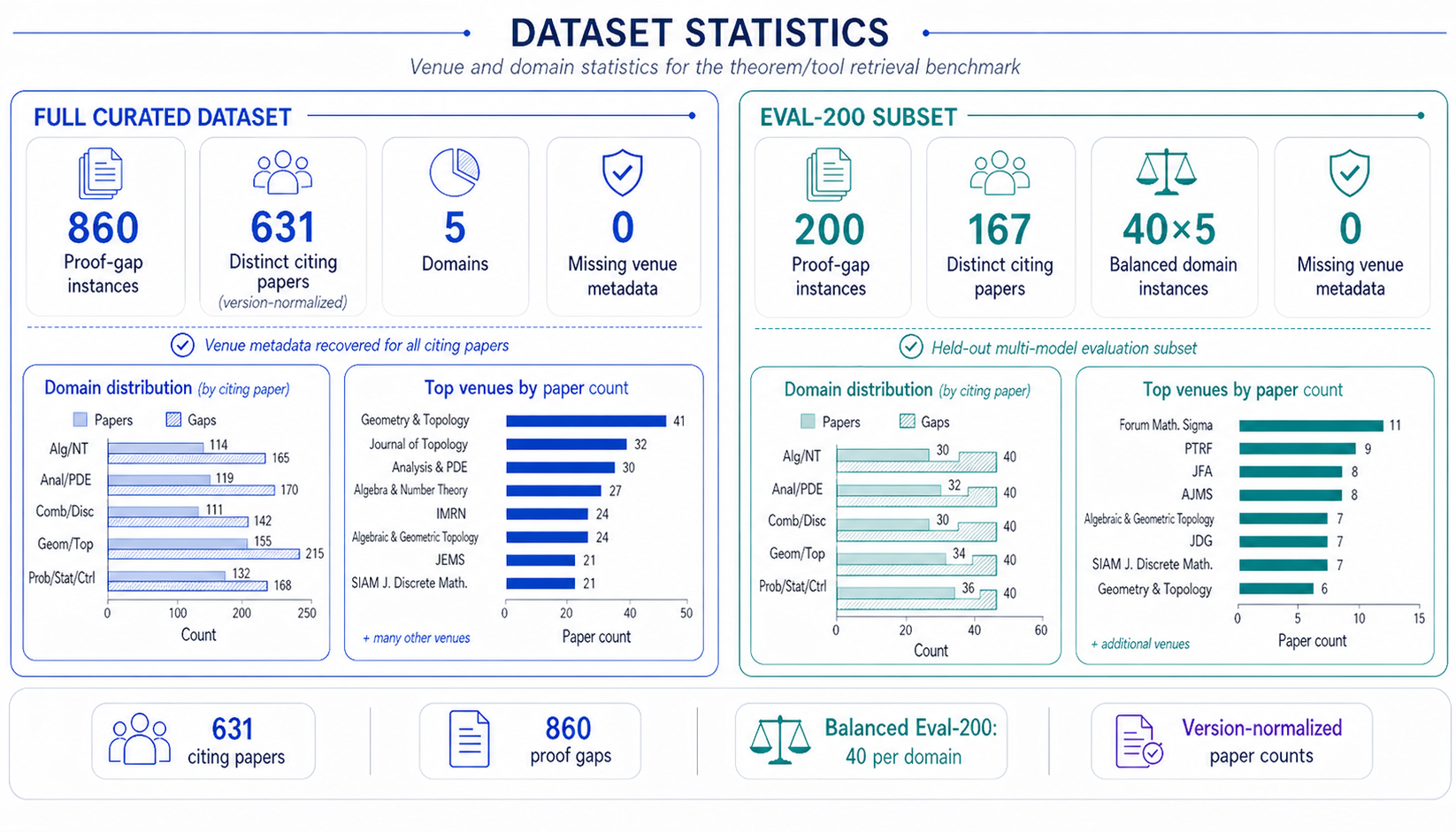}
    \caption{
    \textbf{Dataset statistics.}
    Overview of the curated benchmark and the Eval-200 subset.
    The full curated dataset contains 860 proof-gap instances from 631 version-normalized citing papers.
    Eval-200 contains 200 held-out proof-gap instances from 167 citing papers, balanced across five domains.
    }
    \label{fig:dataset_statistics}
\end{figure}

The main text reports compact domain coverage.  Figure~\ref{fig:dataset_statistics} gives the full release summary used for sanity checks and reproducibility.

\section{Evaluation Protocol and Prompt Templates}
\label{app:evaluation_details}

This section gives the implementation of the abstract retrieval operator and validity predicates used in the main text.  The primary condition disables model-native web-search, browser, file-search, and deep-research tools.  Models may generate queries, but query execution and candidate retrieval are supplied by the frozen Google Scholar retrieval artifact.

\subsection{Input Rendering and Retrieval Artifact}
\label{subsec:appendix_rendering_retrieval}

\paragraph{Input rendering.}
The evaluator renders inputs directly from canonical rows. On the Raw track, a model receives the global setup, target theorem, and ordered local proof blocks. On the Assisted track, the same context is shown with the stored sanitized anchor hint. The canonical end-to-end condition uses the Raw track with the five-block local window.

\paragraph{Why Google Scholar.}
The paper-facing retrieval artifact is instantiated with Google Scholar because the task targets open-world mathematical tool discovery across heterogeneous scholarly sources. The needed tool may appear in a published article, an arXiv or other preprint record, a proceedings paper, a book, a survey, or a bibliographic record that links to a resolvable source. An arXiv-only index would make construction easier to reproduce but would bias evaluation toward recent and source-available papers; a publisher-specific index would miss many preprints, books, and older mathematical records. Google Scholar provides a broad scholarly-search interface across disciplines and source types. We therefore use it as a standardized discovery layer, while freezing its observed outputs because its internal corpus coverage and ranking function are not fully specified by us.

\paragraph{Retrieval execution and temporal controls.}
The paper-facing retrieval artifact is a frozen Google Scholar query cache. It is not defined by a common start date. Instead, retrieval uses a per-instance bibliographic cutoff. For each instance \(i\), the evaluator runs model-authored queries with no lower-bound year and with an upper-bound year equal to the publication year of the original citing paper. This policy is intended to prevent later literature from being used to solve a proof gap from an earlier paper, while preserving access to older papers, books, surveys, and preprints. Because Google Scholar filtering is year-level, this is a publication-year cutoff rather than a reconstruction of the exact day-level search state available to the original authors.

For each model-authored query, the evaluator executes the query once through Google Scholar under fixed externally controlled settings and caches the returned Top-20 candidate prefix. The fixed settings were:
\[
\begin{array}{ll}
\text{interface/result language} & = \text{English},\\
\text{region or domain} & = \text{not additionally controlled},\\
\text{ranking mode} & = \text{default relevance ranking},\\
\text{date-range policy} & = \text{no lower-bound year; upper year = citing-paper publication year},\\
\text{patent results} & = \text{excluded}.
\end{array}
\]
The cache records the query string, execution timestamp, returned rank order, candidate metadata, the externally controlled Google Scholar settings, and source-materialization outcomes when materialization is attempted. Reported numbers are computed only from this frozen query-result artifact.

\paragraph{Temporal interpretation.}
The bibliographic cutoff and the artifact freeze serve different purposes. The per-instance Google Scholar date filter restricts candidate records to the literature dated no later than the original citing paper's publication year. The artifact freeze makes the observed results reproducible: later changes in Google Scholar coverage, metadata, or ranking are not incorporated into the reported numbers. We do not manually promote the author-cited source, rerank candidates using the stored witness, or expose hidden retrieval signals to models. Accordingly, \(\mathrm{CiteRecall@20}\) should be interpreted as a diagnostic of model query choice interacting with the frozen Google Scholar artifact under this date-filter policy, not as a reconstruction of the exact historical search process of the original authors.

\paragraph{Candidate composition.}
Returned candidates may include journal or proceedings articles, arXiv or other preprint records, books, surveys, publisher pages, and bibliographic records when surfaced by Google Scholar. Candidate type alone does not determine correctness: a source is admissible only if the model's extracted statement is grounded in that source and sufficient for the proof transition. Conversely, the author-cited source is used for construction and diagnostics, but exact recovery of that source is not required for \(\mathrm{ToolAcc}\).

\paragraph{Ranking.}
The rank order used in the benchmark is the order returned by Google Scholar for the cached query under the fixed settings above. We do not manually promote the author-cited source, rerank candidates using the stored witness, or expose hidden retrieval signals to models. Thus \(\rho_N\) in the task definition should be understood as the frozen observed mapping from model-authored queries to cached ranked candidate prefixes under this release artifact.

\paragraph{Source materialization.}
After inspecting Top-20 metadata, the model may select at most one candidate to open. If the selected candidate can be resolved under the release protocol, the solver receives the fixed text materialization used by all models. This materialization step deliberately avoids comparing model-native PDF readers, OCR systems, equation renderers, or proprietary document-ingestion tools in the primary leaderboard. The source-materialization log records whether a selected or oracle source was resolved successfully.

\paragraph{Interpretation of retrieval diagnostics.}
\(\mathrm{CiteRecall@20}\) asks whether the stored author-cited witness appears among the cached Top-20 candidates under the source-identity matcher in Section~\ref{subsec:appendix_source_matching}. It is therefore a diagnostic of the interaction between model query choice, Google Scholar coverage, Google Scholar ranking, and source-identity matching. It is not a pure model-ability score and is not the official success criterion. The official criterion remains citation-invariant: any admissible grounded source--tool pair can receive credit.

\subsection{Source Identity Matching}
\label{subsec:appendix_source_matching}

Citation-side diagnostics use a source-identity matcher \(\mathsf{Match}(d,\mathsf{z}_i)\) between retrieved candidate metadata and stored citation metadata.  The matcher prioritizes DOI and arXiv identifiers when available; otherwise it applies normalized title--author matching, with ambiguous journal/preprint variants handled by conservative or manual rules.  This matcher is used only for \(\mathrm{CiteRecall@}N\) and oracle-source diagnostics.  It is not the task definition: correctness is determined by \(\mathsf{Ground}\), \(\mathsf{Suff}\), and membership in \(\mathcal V_i\).

\paragraph{Citing-paper exclusion.}
The citing paper itself, including its arXiv and published variants, is not an admissible retrieved source for the target gap merely because it contains the citation context.  A prediction that restates the local proof sentence or points back to the citing proof does not satisfy \(\mathsf{Ground}\).  The only exception is when the selected text independently states or proves the external tool outside the local citation-dependent argument; otherwise the benchmark would collapse into finding the original paper rather than retrieving the invoked scholarly tool.

\subsection{Source Shortlisting and Full-Text Extraction}
\label{subsec:appendix_shortlisting_extraction}

The model does not read all retrieved documents.  It first shortlists from Top-20 metadata, then opens at most one selected source in full text when that source can be materialized.  The solver extracts, or faithfully restates, a theorem-like statement from the selected source.  A prediction can therefore fail at source selection, at grounding, or at sufficiency even when an apparently relevant paper appears in the candidate list.

\subsection{Evaluation Prompt Templates}
\label{subsec:appendix_prompt_templates}

The following prompt contracts are complete with respect to role, input, constraints, and output fields.  Full released prompt strings preserve these contracts while adding instance renderings and formatting delimiters.

\begin{promptbox}{E1. Planning prompt}
\small
\textbf{Role.} Mathematical Research Assistant.

\textbf{Input.} Global context, target theorem, and pre-citation local proof window.

\textbf{Task.} Infer the immediate next proof intention before retrieval.  Describe what external theorem, estimate, criterion, structural result, comparison principle, gluing property, compactness result, or inequality would close the next proof transition.

\textbf{Constraint.} Return a sanitized planning hint.  Do not mention theorem titles, author names, citation keys, source titles, DOI strings, arXiv identifiers, or bibliographic metadata.

\textbf{Output.} JSON with field \texttt{planning\_anchor}.
\end{promptbox}

\subsection{AnchorMatch Rubric and Examples}
\label{subsec:appendix_anchor_match}

\paragraph{Purpose.}
\(\mathrm{AnchorAcc}\) is a Raw-track planning diagnostic, not an end-to-end success metric.  It asks whether the model can infer the immediate proof need before retrieval.  The stored anchor is a sanitized description of the intended proof role of the missing external tool; it is not a theorem title, source identifier, or gold answer.  Accordingly, \(\mathsf{AnchorMatch}\) compares proof intentions rather than bibliographic identities.

\paragraph{Rubric.}
A model-produced planning description matches the stored anchor when it identifies the same local mathematical obligation: the relevant objects, the direction of use, and the type of conclusion needed to continue the proof.  Exact wording is not required.  Benign paraphrases, equivalent terminology, or a different theorem family can match if they describe the same proof role.  By contrast, broad topical overlap is insufficient.  A description that only says ``use compactness,'' ``apply a comparison theorem,'' or ``find a result about positive definite functions'' does not match unless it specifies the objects and the proof obligation tightly enough to guide retrieval.  Bibliographic shortcuts are also insufficient: mentioning an author, theorem name, source title, citation key, DOI, or arXiv identifier does not by itself establish an anchor match.

\paragraph{Relation to sufficiency.}
Anchor matching is weaker than \(\mathsf{Suff}\).  It does not require the model to state a theorem with all hypotheses, and it does not verify that any source actually contains the needed result.  It only evaluates whether the model has formed the right pre-retrieval intention.  A model can therefore have high \(\mathrm{AnchorAcc}\) but low \(\mathrm{ToolAcc}\) if its queries fail, if it selects the wrong source, or if it extracts a grounded but insufficient statement.  Conversely, a model can occasionally fail \(\mathrm{AnchorAcc}\) but still succeed end-to-end by finding a valid alternative route.

\paragraph{Examples.}
A positive match requires the same proof role, not the same wording. For example, if the stored anchor asks for a compactness result that turns uniform bounds and time-translation control into strong convergence in the topology needed for a nonlinear limit passage, then a model description invoking an Aubin--Lions or Rellich-type compactness theorem for that specific convergence role is counted as a match. Similarly, if the stored anchor asks for a local-to-global gluing principle for a group-theoretic construction, then a model description of a patching theorem upgrading locally verified data to the desired global equality is counted as a match.

Negative cases are also important. A response such as ``use harmonic analysis on spheres'' does not match an anchor asking for a characterization of positive definite functions on spheres via expansion coefficients, because it is only topically related. A response asking for a controllability theorem does not match an anchor asking for non-controllability in a specified fractional-heat parameter regime, because the direction of use is wrong. A bibliographic shortcut such as ``find the theorem in the cited paper'' also does not match unless it states the mathematical proof role. Borderline descriptions such as ``use compactness to get a subsequence'' are marked uncertain or negative when they omit the topology, objects, or conclusion needed by the local proof step; only clear positive matches contribute to \(\mathrm{AnchorAcc}\).

\begin{promptbox}{E2. Query-generation prompt}
\small
\textbf{Role.} Mathematical literature search assistant.

\textbf{Input.} Global setup, target theorem, local proof window, and optionally a planning anchor.

\textbf{Task.} Write a concise scholarly search query aimed at finding a theorem, estimate, lemma, criterion, or standard result that can justify the next proof step.  Prefer mathematical objects, canonical terminology, named properties, theorem families, and strength indicators over generic application words.

\textbf{Rules.} Avoid copying unique prose from the citing paper.  Do not use author names, citation keys, DOI strings, arXiv identifiers, or bibliographic metadata from the stored witness.  Keep the query short enough for a keyword-oriented scholarly search backend.

\textbf{Output.} JSON with field \texttt{query}.
\end{promptbox}

\begin{promptbox}{E3. Metadata shortlisting prompt}
\small
\textbf{Role.} Careful mathematical retrieval shortlister.

\textbf{Input.} Proof context, planned proof need, search query, and Top-\(k\) candidate metadata.

\textbf{Task.} Select the single candidate source most worth opening in full text.  Decide from title, authors, venue, snippets, and metadata whether the candidate is likely to contain a theorem-like tool for the proof gap.

\textbf{Rules.} Do not solve the proof from metadata alone.  Do not select the citing paper merely because it contains the citation context.  Prefer sources likely to state or prove the external tool.

\textbf{Output.} JSON with selected candidate index, selected title, and a brief reason.
\end{promptbox}

\begin{promptbox}{E4. Full-text solver prompt}
\small
\textbf{Role.} Mathematical assistant for the gap-filling task.

\textbf{Input.} Benchmark context and the full text of one selected candidate paper, truncated to a fixed prompt budget.

\textbf{Task.} Decide whether the selected paper contains a theorem-like statement that bridges the proof gap.  If so, extract or faithfully restate the minimal relevant theorem, lemma, proposition, corollary, criterion, estimate, or standard result.

\textbf{Rules.} Preserve hypotheses, domains, quantifiers, parameter ranges, and conclusion strength.  Do not return a theorem title alone.  Do not use the local citing proof sentence as the source of the tool.  If the paper does not support a usable statement, abstain.

\textbf{Output.} JSON with \texttt{relevance\_score}, short reason, and \texttt{extracted\_theorem}.
\end{promptbox}

\begin{promptbox}{E5. Grounding-judge prompt}
\small
\textbf{Input.} Selected source identifier, predicted theorem-like statement, and source excerpts selected by overlap with the prediction.

\textbf{Task.} Return \texttt{yes}, \texttt{no}, or \texttt{uncertain} depending on whether the predicted statement is verbatim or a mathematically faithful restatement of a theorem-like statement supported by the selected source.

\textbf{Scope.} Grounding verifies provenance, not independent truth of the source theorem.  A statement can be grounded even if it is not sufficient for the proof gap.

\textbf{Output.} Label, short reason, confidence, and evidence span.
\end{promptbox}

\begin{promptbox}{E6. Sufficiency-judge prompt}
\small
\textbf{Input.} Rendered proof context, reference witness \(K^\star\), and predicted theorem statement \(\hat K\).

\textbf{Task.} Return \texttt{yes}, \texttt{no}, or \texttt{uncertain} depending on whether \(\hat K\), together with the proof context, is sufficient to justify the target proof transition. Use \(K^\star\) as one reference witness for the intended proof role, but do not require exact identity with \(K^\star\). Accept equivalent, stronger, or otherwise admissible alternative statements when their assumptions are established by the context and their conclusions close the same gap.

\textbf{Checks.} Verify objects, domains, hypotheses, parameter ranges, quantifiers, and conclusion strength. Penalize true but weaker, mismatched, or inapplicable statements.

\textbf{Output.} Label, short reason, and confidence.
\end{promptbox}

\begin{promptbox}{E7. Oracle-source extraction prompt}
\small
\textbf{Role.} Mathematical assistant running an oracle-source extraction experiment.

\textbf{Input.} Benchmark context and the materialized author-cited source content.

\textbf{Task.} Because the source is guaranteed to contain the author-used witness when materialized, identify and extract the theorem-like statement that solves the gap.

\textbf{Output.} JSON with short reason and \texttt{extracted\_theorem}.
\end{promptbox}

\subsection{Grounding, Sufficiency, and Oracle Evaluation}
\label{subsec:appendix_judge_oracle}

A held-out judge scores two predicates.  \textsc{Ground} checks whether the predicted theorem-like statement is supported by the selected source; it verifies provenance, not the mathematical truth of the source theorem itself.  \textsc{Suff} checks whether the statement closes the proof gap under the given context.  \textsc{ToolAcc} is their conjunction.  Oracle-source evaluation bypasses open-world retrieval: when the cited source can be resolved and materialized, the model is given that source directly and asked to extract the relevant theorem-like statement.  This diagnostic isolates within-source localization, statement extraction, and applicability checking from candidate acquisition.

\subsection{Oracle Source Materialization}
\label{subsec:oracle_materialization_audit}

Oracle-source evaluation is defined only for instances whose author-cited source can be resolved and materialized under the release protocol.  This coverage is a property of source access, metadata matching, and parsing, not a model-performance metric.  Table~\ref{tab:oracle_materialization_taxonomy} records the 74-instance unified oracle-evaluable subset, the stable materialized subset, and the run-specific materialization differences.  The main oracle table reports the completed oracle runs and should therefore be read as an upper-bound-style diagnostic rather than a perfectly controlled same-subset rerun.

\begin{table*}[t]
\centering
\small
\caption{\textbf{Oracle-source materialization summary.}  Counts describe the release-protocol source-materialization artifact, not model capability.}
\label{tab:oracle_materialization_taxonomy}
\renewcommand{\arraystretch}{1.10}
\setlength{\tabcolsep}{6pt}
\resizebox{\textwidth}{!}{%
\begin{tabular}{@{}p{0.30\textwidth}p{0.16\textwidth}p{0.54\textwidth}@{}}
\toprule
\textbf{Materialization status} & \textbf{Count} & \textbf{Interpretation} \\
\midrule
Unified oracle-evaluable subset & 74/200 & Author-cited source could be resolved and materialized in the release-protocol artifact. \\
Materialized in every completed run & 66/200 & Stable subset shared across all completed oracle runs. \\
Unavailable in every completed run & 126/200 & Source could not be materialized under the current release protocol. \\
Run-specific materialization differences & 8/200 & Cases with materialization/logging differences across completed oracle runs. \\
\bottomrule
\end{tabular}}
\end{table*}

\section{Additional Results and Diagnostics}
\label{app:additional_results}

This section contains diagnostics that support the main results without changing the leaderboard definition: planning and query ablations, uncertainty estimates, domain stratification, single-annotator audit results, and source-level failure decomposition.  Completed metric tables use the seven model runs reported in the main text.  The audit results below are human-confirmed single-annotator checks; they are not an inter-annotator agreement study.

\subsection{Planning-Window and Query Ablations}
\label{subsec:appendix_planning_query}

Planning diagnostics evaluate whether a Raw-track model can infer a sanitized proof intention before retrieval.  Query diagnostics evaluate whether an Assisted-track query targets the intended tool family after the anchor is supplied.  These metrics localize failures before source selection and extraction; they are not part of the primary leaderboard.

\begin{table*}[t]
\centering
\scriptsize
\caption{\textbf{Planning-window and assisted-query ablations.}  Values are percentages with counts in parentheses.  Raw planning is measured under different context renderings; assisted query accuracy uses the sanitized anchor and measures whether the generated query targets the intended tool family.}
\label{tab:appendix_planning_query}
\renewcommand{\arraystretch}{1.10}
\setlength{\tabcolsep}{3.5pt}
\resizebox{\textwidth}{!}{%
\begin{tabular}{@{}lccccc@{}}
\toprule
& \multicolumn{4}{c}{Raw planning: \(\mathrm{AnchorAcc}\)} & \multicolumn{1}{c}{Assisted query} \\
\cmidrule(lr){2-5}\cmidrule(l){6-6}
\textbf{Model} &
\shortstack[c]{local only\\\(m=5\)} &
\shortstack[c]{global+local\\\(m=1\)} &
\shortstack[c]{global+local\\\(m=3\)} &
\shortstack[c]{global+local\\\(m=5\)} &
\shortstack[c]{\(\mathrm{QueryAcc}\)\\\(m=5\)} \\
\midrule
GPT-5.2             & 10.5\% (21/200) & 8.0\% (16/200)  & 6.0\% (12/200)  & 5.5\% (11/200)  & 18.5\% (37/200) \\
Gemini 3.1 Pro      & 6.5\% (13/200)  & 4.0\% (8/200)   & 3.5\% (7/200)   & 2.0\% (4/200)   & 11.5\% (23/200) \\
Claude Opus 4.5     & 23.0\% (46/200) & 30.0\% (60/200) & 23.5\% (47/200) & 24.0\% (48/200) & 41.0\% (82/200) \\
DeepSeek V3.2       & 11.5\% (23/200) & 7.0\% (14/200)  & 10.5\% (21/200) & 8.0\% (16/200)  & 14.0\% (28/200) \\
Qwen3-235B Thinking & 41.5\% (83/200) & 41.5\% (83/200) & 33.0\% (66/200) & 40.5\% (81/200) & 30.0\% (60/200) \\
Kimi K2 Thinking    & 44.5\% (89/200) & 44.5\% (89/200) & \textbf{52.0\%} (104/200) & 43.5\% (87/200) & 48.0\% (96/200) \\
Grok 4              & 24.4\% (48/197) & 27.0\% (54/200) & 27.5\% (55/200) & 26.5\% (53/200) & \textbf{55.6\%} (110/198) \\
\bottomrule
\end{tabular}}
\begin{minipage}{0.98\textwidth}
\footnotesize
\emph{Note.} Denominators differ for two Grok diagnostic cells because a small number of diagnostic judgments were unavailable.
\end{minipage}
\end{table*}

\paragraph{Interpretation.}
The planning ablation is context-sensitive but not monotone.  GPT-5.2 and Gemini score higher with local-only context than with global+local \(m=5\), suggesting that global setup can distract from the immediate proof transition.  Claude peaks at global+local \(m=1\), Kimi peaks at \(m=3\), Qwen remains high across local-only and \(m=5\), and Grok is comparatively stable.  The assisted-query diagnostic exposes a later bottleneck: Grok, Kimi, Claude, and Qwen reach \(55.6\%\), \(48.0\%\), \(41.0\%\), and \(30.0\%\) QueryAcc, but their end-to-end \(\mathrm{ToolAcc}\) values remain \(3.5\%\), \(3.5\%\), \(7.0\%\), and \(1.0\%\).  Thus an on-target search intention is not sufficient for selecting, reading, and applying the right mathematical tool.  These diagnostics explain upstream errors; they do not replace \(\mathrm{ToolAcc}\), which scores the final source--tool pair.
Because the construction audit identifies residual anchor-sanitization failures, Assisted \(\mathrm{QueryAcc}\)
should be interpreted as a diagnostic under imperfect anchors rather than as a leaderboard metric.
The Raw end-to-end \(\mathrm{ToolAcc}\) condition never exposes anchor text, and therefore is not directly
affected by anchor leakage.
\subsection{Statistical Uncertainty}
\label{subsec:appendix_uncertainty}

Because Eval-200 contains multiple gaps from some citing papers, we report both ordinary binomial intervals and paper-cluster bootstrap intervals.  Table~\ref{tab:appendix_toolacc_ci} gives uncertainty estimates for \(\mathrm{ToolAcc}\).  Full metric-level intervals are included in the aggregation artifact.

\begin{table*}[t]
\centering
\small
\caption{\textbf{Uncertainty for \(\mathrm{ToolAcc}\).}  Wilson intervals use instance-level binomial uncertainty; paper-cluster intervals resample citing papers.}
\label{tab:appendix_toolacc_ci}
\renewcommand{\arraystretch}{1.08}
\setlength{\tabcolsep}{5pt}
\resizebox{\textwidth}{!}{%
\begin{tabular}{@{}lccc@{}}
\toprule
\textbf{Model} & \(\mathrm{ToolAcc}\) & Wilson 95\% CI & Paper-cluster 95\% CI \\
\midrule
Claude Opus 4.5      & 7.0\% (14/200) & [4.2\%, 11.4\%] & [3.6\%, 10.7\%] \\
Grok 4               & 3.5\% (7/200)  & [1.7\%, 7.0\%]  & [1.0\%, 6.6\%] \\
Kimi K2 Thinking     & 3.5\% (7/200)  & [1.7\%, 7.0\%]  & [1.0\%, 6.4\%] \\
GPT-5.2              & 3.0\% (6/200)  & [1.4\%, 6.4\%]  & [1.0\%, 5.5\%] \\
DeepSeek V3.2        & 2.5\% (5/200)  & [1.1\%, 5.7\%]  & [0.5\%, 5.0\%] \\
Gemini 3.1 Pro       & 2.0\% (4/200)  & [0.8\%, 5.0\%]  & [0.5\%, 4.0\%] \\
Qwen3-235B Thinking  & 1.0\% (2/200)  & [0.3\%, 3.6\%]  & [0.0\%, 2.6\%] \\
\bottomrule
\end{tabular}}
\end{table*}

\subsection{Domain Stratification}
\label{subsec:appendix_domain_stratification}

Eval-200 is balanced by domain, with 40 instances in each of five broad areas.  Table~\ref{tab:appendix_domain_toolacc} reports \(\mathrm{ToolAcc}\) by domain.  Because each cell contains only 40 instances, the table should be read as a diagnostic view rather than a stable domain ranking.

\begin{table*}[t]
\centering
\scriptsize
\caption{\textbf{Domain-level \(\mathrm{ToolAcc}\) on Eval-200.}  Each domain has 40 instances.}
\label{tab:appendix_domain_toolacc}
\renewcommand{\arraystretch}{1.08}
\setlength{\tabcolsep}{3pt}
\resizebox{\textwidth}{!}{%
\begin{tabular}{@{}lccccc@{}}
\toprule
\textbf{Model} & \textbf{Alg./NT} & \textbf{Anal./PDE} & \textbf{Comb.} & \textbf{Geom./Top.} & \textbf{Prob./Stat./Ctrl.} \\
\midrule
GPT-5.2             & 0.0\% & 5.0\% & 5.0\%  & 2.5\% & 2.5\% \\
Gemini 3.1 Pro      & 0.0\% & 5.0\% & 0.0\%  & 2.5\% & 2.5\% \\
Claude Opus 4.5     & 5.0\% & 2.5\% & 10.0\% & 7.5\% & 10.0\% \\
DeepSeek V3.2       & 0.0\% & 0.0\% & 5.0\%  & 0.0\% & 7.5\% \\
Qwen3-235B Thinking & 0.0\% & 0.0\% & 0.0\%  & 2.5\% & 2.5\% \\
Kimi K2 Thinking    & 0.0\% & 0.0\% & 7.5\%  & 2.5\% & 7.5\% \\
Grok 4              & 0.0\% & 5.0\% & 5.0\%  & 2.5\% & 5.0\% \\
\bottomrule
\end{tabular}}
\end{table*}

\subsection{Single-Annotator Judge Audit and Source-Level Failure Decomposition}
\label{subsec:appendix_judge_failure}

The main metrics depend on the \textsc{Ground} and \textsc{Suff} predicates, so we audited a stratified sample of judge outputs.  A single human annotator confirmed assistant-prepared signoff judgments on 2026-04-26.  This audit is not an inter-annotator agreement study: the reported agreement and \(\kappa\) compare the fixed judge labels against the single human-confirmed audit labels.

\begin{table*}[t]
\centering
\small
\caption{\textbf{Single-annotator audit of judge labels.}  The audit covers 175 sampled outputs stratified by model and predicted outcome.  Eligible labels exclude abstentions and cases without an evaluable prediction; here abstentions are recorded as \texttt{uncertain} by both the fixed judge and the human-confirmed audit label.  \(\kappa\) is judge-vs-human-confirmed kappa, not inter-annotator kappa.}
\label{tab:judge_human_audit}
\renewcommand{\arraystretch}{1.10}
\setlength{\tabcolsep}{5pt}
\resizebox{\textwidth}{!}{%
\begin{tabular}{@{}lcccccc@{}}
\toprule
\textbf{Predicate} & \textbf{Eligible} & \textbf{Agreement} & \textbf{Precision} & \textbf{Recall} & \textbf{F1} & \(\boldsymbol{\kappa}\) \\
\midrule
\textsc{Ground}  & 156 & 100.0\% & 100.0\% & 100.0\% & 100.0\% & 1.000 \\
\textsc{Suff}    & 156 & 98.1\%  & 93.3\%  & 100.0\% & 96.6\%  & 0.952 \\
\textsc{ToolAcc} & 155 & 98.1\%  & 93.3\%  & 100.0\% & 96.6\%  & 0.952 \\
\bottomrule
\end{tabular}}
\end{table*}

\begin{table*}[t]
\centering
\small
\caption{\textbf{Judge-audit sample composition.}  The 175 audited outputs are stratified across success, grounded-but-insufficient, ungrounded, and judge-abstained buckets as well as across the seven reported models.}
\label{tab:judge_audit_composition}
\renewcommand{\arraystretch}{1.10}
\setlength{\tabcolsep}{5pt}
\resizebox{\textwidth}{!}{%
\begin{tabular}{@{}lrrrrrrr r@{}}
\toprule
\textbf{Bucket} & \textbf{GPT} & \textbf{Gemini} & \textbf{Claude} & \textbf{DeepSeek} & \textbf{Qwen} & \textbf{Kimi} & \textbf{Grok} & \textbf{Total} \\
\midrule
Success & 6 & 4 & 14 & 5 & 2 & 7 & 7 & 45 \\
Grounded but insufficient & 10 & 4 & 8 & 12 & 9 & 9 & 18 & 70 \\
Ungrounded & 5 & 7 & 6 & 5 & 8 & 5 & 4 & 40 \\
Judge abstained & 3 & 3 & 2 & 3 & 4 & 3 & 2 & 20 \\
\midrule
Total & 24 & 18 & 30 & 25 & 23 & 24 & 31 & 175 \\
\bottomrule
\end{tabular}}
\end{table*}

Alternative-source successes are central to the citation-invariant claim, so the audit also checks their admissibility.  The human-confirmed audit validates 39 of 42 judge-scored alternative-source successes and all 3 cited-source successes.  The three invalid alternative-source successes are false positives for sufficiency or source admissibility; Table~\ref{tab:alt_source_invalid_cases} records their impact as an audit sensitivity analysis.  The main reported aggregate tables remain the fixed-judge benchmark scores, while this audit quantifies their residual label risk.

\begin{table*}[t]
\centering
\small
\caption{\textbf{Human audit of successful source type.}  The audit covers all judge-scored successful outputs in Eval-200.}
\label{tab:alt_source_human_audit}
\renewcommand{\arraystretch}{1.10}
\setlength{\tabcolsep}{10pt}
\begin{tabular}{@{}lcc@{}}
\toprule
\textbf{Success type} & \textbf{Valid / audited} & \textbf{Validity} \\
\midrule
Alternative-source successes & 39 / 42 & 92.9\% \\
Cited-source successes & 3 / 3 & 100.0\% \\
\bottomrule
\end{tabular}
\end{table*}

\begin{table*}[t]
\centering
\scriptsize
\caption{\textbf{Audit-flagged invalid alternative-source successes.}  These cases are useful for sensitivity analysis because they show how much the fixed-judge ToolAcc would move if the single-annotator audit labels were applied retroactively.}
\label{tab:alt_source_invalid_cases}
\renewcommand{\arraystretch}{1.10}
\setlength{\tabcolsep}{3pt}
\resizebox{\textwidth}{!}{%
\begin{tabular}{@{}p{0.08\textwidth}p{0.17\textwidth}p{0.55\textwidth}p{0.15\textwidth}@{}}
\toprule
\textbf{Model} & \textbf{Instance} & \textbf{Audit reason} & \textbf{If corrected} \\
\midrule
GPT-5.2 & \texttt{2012.12831\_gap\_0} & Recovered the Boolean-to-tropical transfer theorem, but not the monotone lower bound needed to instantiate it. & ToolAcc \(-1\); AltSourceToolAcc \(-1\) \\
GPT-5.2 & \texttt{2401.14380v1\_gap\_0} & Retrieved the paper's own target theorem rather than the external spline-formula witness used for the proof step. & ToolAcc \(-1\); AltSourceToolAcc \(-1\) \\
Claude Opus 4.5 & \texttt{2010.01779v3\_gap\_0} & Retrieved only the heavy-tail trap assumption; the audited witness is a scaling-limit theorem. & ToolAcc \(-1\); AltSourceToolAcc \(-1\) \\
\bottomrule
\end{tabular}}
\end{table*}

Two additional cases were protocol flags rather than predicate disagreements: \texttt{1909.11981v3\_gap\_3} for Claude and \texttt{2109.07887v3\_gap\_1} for Qwen selected sources matching the citing paper or its title.  These flags motivate the explicit citing-paper exclusion rule in Appendix~\ref{subsec:appendix_source_matching} and should be revisited in any future audit-corrected scoring artifact.

The paper-facing failure decomposition is source-level rather than citation-level.  Because the task is citation-invariant, failures are not defined by whether the author-cited source appears in Top-20.  Table~\ref{tab:appendix_failure_decomp} partitions final outcomes into no grounded support, grounded but insufficient, and success.  Retrieval-conditioned views based on \(\mathrm{CiteRecall@20}\) remain useful for debugging, but they are not the official failure definition.

\begin{table*}[t]
\centering
\small
\caption{\textbf{Source-level failure decomposition on Eval-200.}  ``Grounded but insufficient'' equals \(\mathrm{GroundRate}-\mathrm{ToolAcc}\) under the fixed-judge benchmark scores.}
\label{tab:appendix_failure_decomp}
\renewcommand{\arraystretch}{1.08}
\setlength{\tabcolsep}{5pt}
\resizebox{\textwidth}{!}{%
\begin{tabular}{@{}lccc@{}}
\toprule
\textbf{Model} & \textbf{No grounded support} & \textbf{Grounded but insufficient} & \textbf{Success} \\
\midrule
GPT-5.2             & 150/200 (75.0\%) & 44/200 (22.0\%) & 6/200 (3.0\%) \\
Gemini 3.1 Pro      & 176/200 (88.0\%) & 20/200 (10.0\%) & 4/200 (2.0\%) \\
Claude Opus 4.5     & 152/200 (76.0\%) & 34/200 (17.0\%) & 14/200 (7.0\%) \\
DeepSeek V3.2       & 142/200 (71.0\%) & 53/200 (26.5\%) & 5/200 (2.5\%) \\
Qwen3-235B Thinking & 166/200 (83.0\%) & 32/200 (16.0\%) & 2/200 (1.0\%) \\
Kimi K2 Thinking    & 152/200 (76.0\%) & 41/200 (20.5\%) & 7/200 (3.5\%) \\
Grok 4              & 117/200 (58.5\%) & 76/200 (38.0\%) & 7/200 (3.5\%) \\
\bottomrule
\end{tabular}}
\end{table*}

\section{Controlled Retrieval vs. Live Web-Browsing Agents}
\label{app:deep_research_exclusion}

Recent ``deep research'' systems perform long-horizon web research by planning searches, issuing iterative queries, reading many pages or files, and synthesizing cited reports \cite{openai2025deepresearch,google2026geminideepresearch}.  Academic benchmarks increasingly study such agents as report writers or general web researchers \cite{bosse2025deepresearchbench,li2025reportbench}.  We view these systems as complementary to \ourbench, but outside the primary fixed-budget, fixed-backend leaderboard.

\paragraph{Rationale.}
Unrestricted browsing introduces variables that are orthogonal to the mathematical policy under evaluation: live-search drift, proprietary indices, ranking and crawling policy, page-opening strategy, context compression, retry budget, freshness filtering, and provider-specific safety layers.  These hidden operations overlap with the benchmarked stages \(\pi_{\mathrm{qry}},\pi_{\mathrm{sel}},\pi_{\mathrm{ext}}\).  Allowing each model to use a different integrated search stack would therefore compare model-plus-backend systems rather than the controlled source--tool retrieval policy.

\paragraph{Benchmark contract.}
The primary protocol follows the information-retrieval practice of shared test collections and standardized evaluation conditions \cite{sanderson2010testcollections}: agents see the same instances, candidate budget, cached query outputs, and verification policy.  Models still decide what to search for and how to use retrieved evidence, but query execution, candidate ranking, source materialization, context budget, and logs are supplied by the release-fixed backend.

\paragraph{Agentic-learning view.}
The same interface can be viewed as a finite-horizon agentic decision process.  The state contains the rendered proof context, the model's current planning description, previous queries, cached retrieval results, and, after a source-opening action, the fixed source materialization.  The action space consists of planning, query formulation, metadata-based source selection, optional source opening, extraction or faithful restatement of a theorem-like statement, and abstention.  The terminal reward is the source--tool success indicator
\[
\mathbf 1[
\mathsf{Ground}_i(\hat d_i,\hat K_i)
\wedge
\mathsf{Suff}_i(\hat K_i\mid \mathcal C_i,\Delta_i)
],
\]
with auxiliary diagnostics such as \(\mathrm{AnchorAcc}\), \(\mathrm{CiteRecall@20}\), and \(\mathrm{GroundRate}\) available as structured feedback.  Because query results and source materializations are frozen, this environment supports reproducible offline policy analysis and future agentic RL studies without conflating policy learning with live-search drift or proprietary browsing stacks.  The present paper uses this structure for evaluation only; it does not train policies on \ourbench{}.

\paragraph{Citation-derived shortcut control.}
Because instances are mined from explicit proof citations, unrestricted browsing can reward bibliographic shortcuts such as phrase matching, citation-chain following, or direct recovery of the author-cited paper.  Such behavior may be useful in deployed research assistants, but it changes the benchmark target from identifying a sufficient mathematical tool to open-web source hunting.

\paragraph{Future browsing track.}
A future controlled browsing or native-search track should use the same frozen release snapshot, or a separately released browsing snapshot, and should report query budgets, visited pages, opened files, backend details, context-construction rules, and complete interaction logs alongside \textsc{ToolAcc}.

\subsection{Implementation Notes}
\label{app:script_corrections}

Several helper-script options are not part of the official benchmark definition. Solver-as-judge fallback is only a debugging convenience; all paper-facing runs use the fixed held-out judge. Legacy proxy fields such as \texttt{tool\_success\_proxy}, \texttt{is\_matched}, and
\texttt{cited\_tool\_success\_proxy} are not paper metrics. The official retrieval date policy is the Google Scholar year-level cutoff specified in Appendix~\ref{subsec:appendix_rendering_retrieval}.

\section{Detailed Case Studies}
\label{app:case_studies}

This appendix gives a mechanism-level reading of representative instances from the completed seven-model evaluation.  The cases are not an additional split and should not be read as anecdotal successes or failures; they are qualitative probes of the aggregate pattern in Section~\ref{sec:analysis}.  Each case first presents the proof situation and reference witness, then reports the observed model behavior, and finally explains which interface breaks: proof need \(\rightarrow\) search-effective language, retrieved source \(\rightarrow\) theorem-bearing passage, or sourced statement \(\rightarrow\) assumption-checked proof tool.

\paragraph{Legend.}
\newcommand{\caseY}{\textsc{Y}}
\newcommand{\caseN}{\textsc{N}}
\newcommand{\caseNA}{\(\bot\)}
\newcommand{\caseT}{\textsc{T}}
\newcommand{\caseF}{\textsc{F}}
\newcommand{\casena}{--}
\textsc{Y}/\textsc{N} indicate positive/negative labels, and \(\bot\) indicates no evaluable output.  \textbf{Alt} means that a successful source was not the stored author-cited source.  \textbf{Plan} is the five-block Raw-track planning match.  \textbf{Query} is the assisted-query diagnostic.  Titles are shortened for readability.

\begin{table*}[t]
\centering
\footnotesize
\renewcommand{\arraystretch}{1.18}
\setlength{\tabcolsep}{4pt}
\caption{\textbf{Case-study roadmap.}
Each case isolates a distinct mechanism behind low end-to-end \(\mathrm{ToolAcc}\).}
\label{tab:case_summary}

\begin{tabularx}{\textwidth}{
@{}
>{\RaggedRight\arraybackslash}p{0.18\textwidth}
>{\RaggedRight\arraybackslash}p{0.20\textwidth}
>{\RaggedRight\arraybackslash}p{0.27\textwidth}
>{\RaggedRight\arraybackslash}X
@{}}
\toprule
\textbf{Case} & \textbf{Mechanism} & \textbf{Reference tool} & \textbf{Diagnostic readout} \\
\midrule

\makecell[l]{\texttt{1804.05967v2}\\[-0.15em]\texttt{\_gap\_0}}
&
Robust alternative-source success
&
Conditioned planar Brownian motion equals an \(h\)-transform
&
Mathematical validity is not bibliographic identity: several systems close the gap while missing the author citation.
\\
\addlinespace[0.25em]

\makecell[l]{\texttt{1901.06096v3}\\[-0.15em]\texttt{\_gap\_1}}
&
Citation-hit classical theorem success
&
Schoenberg characterization of positive definite functions on spheres
&
Citation recall can be high while theorem-level localization and formulation remain fragile.
\\
\addlinespace[0.25em]

\makecell[l]{\texttt{1510.08891}\\[-0.15em]\texttt{\_gap\_1}}
&
Oracle rescue / retrieval bottleneck
&
Matroidal locus equals the Voronoi--perfect-cone intersection
&
The proof vocabulary points toward cubic threefolds, but the needed tool lives in toroidal-fan language.
\\
\addlinespace[0.25em]

\makecell[l]{\texttt{1804.10581v2}\\[-0.15em]\texttt{\_gap\_12}}
&
Stronger theorem succeeds; nearby theorem fails
&
Non-controllability of fractional heat equation
&
A correct topical source is not enough; the equation, domain, parameter range, and quantifiers must match.
\\
\addlinespace[0.25em]

\makecell[l]{\texttt{1909.12545}\\[-0.15em]\texttt{\_gap\_1}}
&
Grounded but insufficient local corollary
&
Singular \(\mathbb Q\)-factorial terminal varieties admit no proper crepant resolution
&
Source-supported corollaries fail when their extra hypotheses are not established by the proof state.
\\
\addlinespace[0.25em]

\makecell[l]{\texttt{2507.02263v2}\\[-0.15em]\texttt{\_gap\_1}}
&
Weaker theorem under same assumptions
&
Signless-Laplacian excess gives \(\Omega(n^{k+1})\) cliques
&
The same assumptions can support multiple conclusions; only the right strength closes the gap.
\\
\addlinespace[0.25em]

\makecell[l]{\texttt{1905.05370}\\[-0.15em]\texttt{\_gap\_0}}
&
Anchor consensus, query collapse
&
Compactness/convergence theorem for Muskat approximations
&
Models recognize compactness but compress the proof obligation into overly generic method words.
\\
\addlinespace[0.25em]

\makecell[l]{\texttt{2012.12831}\\[-0.15em]\texttt{\_gap\_0}}
&
Semantically plausible query, retrieval/protocol ambiguity
&
Monotone lower bound for the logical permanent
&
Internal transfer principles can look useful while missing the external theorem that instantiates them.
\\
\addlinespace[0.25em]

\makecell[l]{\texttt{2110.11087v2}\\[-0.15em]\texttt{\_gap\_1}}
&
Canonical algebra query, zero retrieval
&
Polynomial gluing for Steinberg groups
&
Even nearly canonical theorem-family language may not surface an older, poorly indexed algebra source.
\\

\bottomrule
\end{tabularx}
\end{table*}

\subsection{Robust Alternative-Source Success: Brownian Random Interlacements}
\label{app:case_brownian_alt_source}

\noindent\fbox{%
\begin{minipage}{0.96\linewidth}
\small
\textbf{Instance.} \texttt{1804.05967v2\_gap\_0}.\quad
\textbf{Paper.} \emph{Two-dimensional Brownian random interlacements}.\quad
\textbf{Domain.} Probability / statistics / control.\\[0.2em]
\textbf{Proof situation.} The proof uses a comparison between planar Brownian motion conditioned to hit a larger circle before a smaller one and a Doob \(h\)-transformed process.\\[0.2em]
\textbf{Reference witness.} For all \(R>1\) and \(1<\|x\|<R\),
\[
\mathbb P_x\!\big[W|_{[0,\tau(R)]}\in\cdot\mid \tau(R)<\tau(1)\big]
=
\mathbb P_x\!\big[\widehat W|_{[0,\widehat\tau(R)]}\in\cdot\big].
\]
\textbf{Construction citation.} Comets--Popov--Vachkovskaia, \emph{Two-dimensional random interlacements and late points for random walks}.
\end{minipage}}

\paragraph{Model behavior.}
\begin{center}
\scriptsize
\renewcommand{\arraystretch}{1.08}
\setlength{\tabcolsep}{3pt}
\resizebox{\linewidth}{!}{%
\begin{tabular}{@{}p{0.18\linewidth}p{0.26\linewidth}ccccccp{0.30\linewidth}@{}}
\toprule
\textbf{Model} & \textbf{Query} & \textbf{Plan} & \textbf{Query} & \textbf{Cite@20} & \textbf{Ground} & \textbf{Suff} & \textbf{Tool} & \textbf{Selected source} \\
\midrule
GPT-5.2 & Brownian sausage covering & \caseN & \caseN & \caseN & \caseN & \caseN & \caseN & LIL for cover times by Wiener sausage \\
Gemini 3.1 Pro & boundedness vacant set Brownian interlacements & \caseN & \caseN & \caseN & \caseN & \caseN & \caseN & Two-Dimensional Brownian Random Interlacements \\
Claude Opus 4.5 & random interlacement covering probability asymptotics & \caseN & \caseN & \caseN & \caseY & \caseY & \caseY & Two-Dimensional Brownian Random Interlacements \\
DeepSeek V3.2 & BRI sausages covering probability limit & \caseN & \caseN & \caseN & \caseY & \caseY & \caseY & Two-Dimensional Brownian Random Interlacements \\
Qwen3-235B Thinking & BRI sausages compact covering probability & \caseY & \caseN & \caseN & \caseY & \caseY & \caseY & Two-Dimensional Brownian Random Interlacements \\
Kimi K2 Thinking & \(h\)-transform transition kernel Brownian & \caseY & \caseN & \caseN & \caseY & \caseY & \caseY & Two-dimensional Brownian random interlacements \\
Grok 4 & BRI covering probability alpha infinity & \caseN & \caseN & \caseN & \caseY & \caseY & \caseY & Two-Dimensional Brownian Random Interlacements \\
\bottomrule
\end{tabular}}
\end{center}

\paragraph{Representative extraction.}
Claude, DeepSeek, Qwen, Kimi, and Grok all recover a statement equivalent to Lemma~2.1 of the selected source: Brownian motion stopped at \(\tau(R)\), conditioned on \(\tau(R)<\tau(1)\), has the same law as the \(h\)-transformed process stopped at \(\widehat\tau(R)\).  The notation varies, but the assumptions and conclusion match the reference witness.

\paragraph{Mechanistic diagnosis.}
This case isolates why citation-invariance is a mathematical requirement rather than an evaluation convenience.  All successful systems have \(\mathrm{CiteRecall@20}=0\), yet their selected source states precisely the path-law identity needed by the proof.  Exact citation recovery would therefore penalize valid scholarly justifications.  At the same time, the upstream planning and query signals are mostly negative, showing that the final object must be judged as a source--tool pair rather than as a sequence of intermediate proxy labels.

\paragraph{Takeaway.}
The lesson is that provenance and applicability are separable from citation identity.  The author citation remains useful for construction and debugging, but the success criterion must ask whether the returned source supports a sufficient tool.  This is exactly the distinction between \(\mathrm{CiteRecall@20}\) as a diagnostic and \(\mathrm{ToolAcc}\) as the task metric.

\subsection{Citation-Hit Classical Theorem Success: Positive Definite Functions on Spheres}
\label{app:case_schoenberg_success}

\noindent\fbox{%
\begin{minipage}{0.96\linewidth}
\small
\textbf{Instance.} \texttt{1901.06096v3\_gap\_1}.\quad
\textbf{Paper.} \emph{Repeated minimizers of \(p\)-frame energies}.\quad
\textbf{Domain.} Combinatorics / discrete mathematics.\\[0.2em]
\textbf{Proof situation.} The proof reduces a \(p\)-frame energy bound to the case \(p=2\), where it invokes positive definite functions on the unit sphere.\\[0.2em]
\textbf{Reference witness.} A continuous function \(f:[-1,1]\to\mathbb R\) is positive definite on \(\mathbb S^{d-1}\) iff it admits a Gegenbauer expansion \(f(t)=\sum_{k\ge0}a_k C_k^{(d-2)/2}(t)\) with \(a_k\ge0\).\\[0.2em]
\textbf{Construction citation.} Schoenberg, \emph{Positive definite functions on spheres}, 1941.
\end{minipage}}

\paragraph{Model behavior.}
\begin{center}
\scriptsize
\renewcommand{\arraystretch}{1.08}
\setlength{\tabcolsep}{3pt}
\resizebox{\linewidth}{!}{%
\begin{tabular}{@{}p{0.18\linewidth}p{0.28\linewidth}ccccccp{0.27\linewidth}@{}}
\toprule
\textbf{Model} & \textbf{Query} & \textbf{Plan} & \textbf{Query} & \textbf{Cite@20} & \textbf{Ground} & \textbf{Suff} & \textbf{Tool} & \textbf{Selected source} \\
\midrule
GPT-5.2 & positive definite polynomial sphere & \caseN & \caseY & \caseY & \caseNA & \caseNA & \caseNA & Strictly positive definite functions on spheres \\
Gemini 3.1 Pro & positive definite function coefficients one & \caseN & \caseY & \caseY & \caseN & \caseN & \caseN & Metric spaces and positive definite functions \\
Claude Opus 4.5 & positive definite functions spheres Gegenbauer & \caseY & \caseY & \caseY & \caseY & \caseY & \caseY & Positive definite functions in distance geometry \\
DeepSeek V3.2 & positive definite functions unit sphere & \caseY & \caseY & \caseY & \caseNA & \caseNA & \caseNA & none \\
Qwen3-235B Thinking & spherical positive definite sum squares & \caseY & \caseN & \caseY & \caseY & \caseN & \caseN & Positive definite functions on the unit sphere and integrals of Jacobi polynomials \\
Kimi K2 Thinking & positive definite sphere quadratic forms & \caseN & \caseY & \caseY & \caseN & \caseN & \caseN & Optimal measures for \(p\)-frame energies \\
Grok 4 & positive definite functions Gegenbauer expansion & \caseY & \caseY & \caseY & \caseN & \caseN & \caseN & Optimal measures for \(p\)-frame energies \\
\bottomrule
\end{tabular}}
\end{center}

\paragraph{Representative success.}
Claude extracts a Schoenberg-type theorem: a continuous real function on the sphere is positive definite iff it is a nonnegative linear combination of Gegenbauer polynomials.  The source uses slightly different notation, but the statement is mathematically equivalent to the witness and directly supplies the positive-definiteness tool for the \(p=2\) energy argument.

\paragraph{Representative near miss.}
Qwen selects a related paper and grounds a true sufficient-condition theorem for positive definiteness involving smoothness and derivative sign conditions.  The theorem is real, but it is not the Schoenberg equivalence and it imposes hypotheses not established in the proof context.  It therefore fails sufficiency.

\paragraph{Mechanistic diagnosis.}
This is a relatively favorable retrieval case: the intended theorem family is classical, most queries contain the right mathematical vocabulary, and the stored citation often appears in the candidate pool.  The difficulty therefore shifts downstream.  The model must recognize that the proof needs Schoenberg's equivalence theorem, not merely any positive-definiteness criterion or a paper about spherical kernels.  Only Claude turns the retrieved literature into a statement with the right iff structure, polynomial basis, and nonnegative-coefficient condition.

\paragraph{Takeaway.}
This case shows that source acquisition is only a coarse gate.  A theorem-use agent must operate below the paper level: it must identify the theorem boundary, preserve the logical form, and decide whether the extracted statement has the right level of generality for the local argument.

\subsection{Oracle Rescue: Matroidal Locus and Cubic Threefold Compactifications}
\label{app:case_oracle_cubic}

\noindent\fbox{%
\begin{minipage}{0.96\linewidth}
\small
\textbf{Instance.} \texttt{1510.08891\_gap\_1}.\quad
\textbf{Paper.} \emph{Complete moduli of cubic threefolds and their intermediate Jacobians}.\quad
\textbf{Domain.} Algebra / number theory.\\[0.2em]
\textbf{Proof situation.} The proof needs a toroidal-compactification fact to show that the extended intermediate-Jacobian map lands in the correct compactification locus.\\[0.2em]
\textbf{Reference witness.} The matroidal locus \(\mathrm{Matr}[5]\) is the maximal partial compactification contained in both the second Voronoi and perfect-cone compactifications; equivalently, \(\Sigma^V\cap\Sigma^P=\Sigma^{\mathrm{mat}}\).\\[0.2em]
\textbf{Construction citation.} Melo--Viviani, \emph{Comparing perfect and 2nd Voronoi decompositions: the matroidal locus}.
\end{minipage}}

\paragraph{Model behavior.}
\begin{center}
\scriptsize
\renewcommand{\arraystretch}{1.08}
\setlength{\tabcolsep}{3pt}
\resizebox{\linewidth}{!}{%
\begin{tabular}{@{}p{0.18\linewidth}p{0.28\linewidth}ccccccp{0.27\linewidth}@{}}
\toprule
\textbf{Model} & \textbf{Query} & \textbf{Plan} & \textbf{Query} & \textbf{Cite@20} & \textbf{Ground} & \textbf{Suff} & \textbf{Tool} & \textbf{Selected source} \\
\midrule
GPT-5.2 & cubic threefold monodromy cones & \caseN & \caseN & \caseN & \caseN & \caseN & \caseN & Moduli space via degenerations \\
Gemini 3.1 Pro & local monodromy boundary divisors & \caseN & \caseN & \caseN & \caseN & \caseN & \caseN & A \(p\)-adic local monodromy theorem \\
Claude Opus 4.5 & monodromy intermediate Jacobian cubic threefold & \caseN & \caseN & \caseN & \caseNA & \caseNA & \caseNA & none \\
DeepSeek V3.2 & monodromy intermediate Jacobian cubic threefolds & \caseN & \caseN & \caseN & \caseN & \caseN & \caseN & Moduli space via degenerations \\
Qwen3-235B Thinking & monodromy GIT boundary Voronoi compatibility & \caseN & \caseN & \caseN & \caseNA & \caseNA & \caseNA & none \\
Kimi K2 Thinking & monodromy intermediate Jacobian GIT boundary & \caseN & \caseN & \caseN & \caseN & \caseN & \caseN & Moduli space via degenerations \\
Grok 4 & monodromy intermediate Jacobian cubic threefolds & \caseN & \caseN & \caseN & \caseY & \caseN & \caseN & Moduli space via degenerations \\
\bottomrule
\end{tabular}}
\end{center}

\paragraph{Representative near miss.}
Grok grounds a statement about the boundary geometry of intermediate Jacobians.  The statement is related to the citing paper, but it is not the fan-theoretic equality \(\Sigma^V\cap\Sigma^P=\Sigma^{\mathrm{mat}}\) and does not imply the maximality of the matroidal partial compactification.

\paragraph{Oracle rescue.}
In the oracle-source condition, GPT and Claude recover the intended statement from Melo--Viviani.  Claude extracts the clean fan identity: a cone belongs to both the second Voronoi and perfect-cone decompositions iff it belongs to the matroidal decomposition.

\paragraph{Mechanistic diagnosis.}
Open retrieval is captured by the surface vocabulary of the citing proof: cubic threefolds, monodromy, degenerations, and intermediate Jacobians.  The external tool, however, is expressed in a different mathematical register: Voronoi and perfect-cone decompositions, fan intersections, and the matroidal locus.  Oracle access changes the problem qualitatively: once Melo--Viviani is supplied, GPT and Claude can recover the relevant fan identity.  The failure is therefore not that the tool is unreadable, but that the model cannot bridge from the local proof vocabulary to the external theorem family.

\paragraph{Takeaway.}
The case demonstrates a central source-acquisition problem: the best search terms for a proof step may not be the nouns appearing in the local proof.  The model must infer the hidden mathematical abstraction that the authors import from another literature.

\subsection{Same Topic, Wrong Theorem: Fractional Heat Non-Controllability}
\label{app:case_fractional_heat}

\noindent\fbox{%
\begin{minipage}{0.96\linewidth}
\small
\textbf{Instance.} \texttt{1804.10581v2\_gap\_12}.\quad
\textbf{Paper.} \emph{Lack of null-controllability for the fractional heat equation and related equations}.\quad
\textbf{Domain.} Probability / statistics / control.\\[0.2em]
\textbf{Proof situation.} The proof needs non-controllability for the fractional heat equation on \(\Omega=\mathbb R\) or \(\mathbb T\), for \(0\le\alpha<1\), on any strict open control subset.\\[0.2em]
\textbf{Reference witness.} If \(0\le\alpha<1\), \(\Omega=\mathbb R\) or \(\mathbb T\), and \(\omega\subsetneq \Omega\) is open, then the fractional heat equation is not null-controllable on \(\omega\) in any time \(T>0\).\\[0.2em]
\textbf{Construction citation.} The citing paper points to an earlier special case for a related phase-space domain and time threshold.
\end{minipage}}

\paragraph{Model behavior.}
\begin{center}
\scriptsize
\renewcommand{\arraystretch}{1.08}
\setlength{\tabcolsep}{3pt}
\resizebox{\linewidth}{!}{%
\begin{tabular}{@{}p{0.18\linewidth}p{0.28\linewidth}ccccccp{0.27\linewidth}@{}}
\toprule
\textbf{Model} & \textbf{Query} & \textbf{Plan} & \textbf{Query} & \textbf{Cite@20} & \textbf{Ground} & \textbf{Suff} & \textbf{Tool} & \textbf{Selected source} \\
\midrule
GPT-5.2 & fractional heat controllability counterexamples & \caseN & \caseN & \caseN & \caseY & \caseN & \caseN & Lack of null-controllability... \\
Gemini 3.1 Pro & fractional heat lack null controllability & \caseN & \caseN & \caseN & \caseY & \caseN & \caseN & Lack of null-controllability... \\
Claude Opus 4.5 & controllability cost blow-up time threshold & \caseN & \caseN & \caseN & \caseNA & \caseNA & \caseNA & none \\
DeepSeek V3.2 & fractional heat null controllability decay & \caseN & \caseN & \caseN & \caseY & \caseY & \caseY & Lack of null-controllability... \\
Qwen3-235B Thinking & high-frequency non-observability fractional heat equation & \caseY & \caseN & \caseN & \caseN & \caseN & \caseN & Se\'minaire de th\'eorie spectrale... \\
Kimi K2 Thinking & fractional heat observability inequality concentration & \caseN & \caseN & \caseN & \caseNA & \caseN & \caseNA & Exact observability properties... \\
Grok 4 & fractional heat equation null-controllability minimal & \caseN & \caseN & \caseN & \caseY & \caseN & \caseN & Lack of null-controllability... \\
\bottomrule
\end{tabular}}
\end{center}

\paragraph{Successful extraction.}
DeepSeek extracts Theorem~1.3 from the selected source: for \(0\le\alpha<1\), \(\Omega=\mathbb R\) or \(\mathbb T\), and strict open \(\omega\), the fractional heat equation is not null-controllable in any time.  This is stronger than the special-case witness and therefore sufficient.

\paragraph{Near misses.}
GPT-5.2, Gemini, and Grok select a topical source and ground real non-controllability or observability statements, but the statements concern a different equation, phase-space domain, or time threshold.  Qwen's query captures high-frequency non-observability but selects an unrelated source.

\paragraph{Mechanistic diagnosis.}
This case makes the sufficiency predicate concrete.  The retrieved papers are not random distractors; they contain genuine observability and non-controllability statements in the same topical neighborhood.  What separates success from failure is exact applicability: the equation class, state space, control set, parameter regime, and quantification over time must all line up with the local proof state.  DeepSeek succeeds because it extracts a statement whose scope covers the target proof obligation, whereas several grounded near misses fail on one of these dimensions.

\paragraph{Takeaway.}
This is the pattern behind the large \(\mathrm{GroundRate}\)-to-\(\mathrm{ToolAcc}\) gap in the main tables.  Grounding answers the provenance question, but the proof requires a second judgment: whether the sourced statement is the right mathematical instrument for this exact transition.

\subsection{Grounded but Insufficient: Symplectic Resolutions and Birational Obstructions}
\label{app:case_symplectic_resolution}

\noindent\fbox{%
\begin{minipage}{0.96\linewidth}
\small
\textbf{Instance.} \texttt{1909.12545\_gap\_1}.\quad
\textbf{Paper.} \emph{Symplectic resolutions of character varieties}.\quad
\textbf{Domain.} Geometry / topology.\\[0.2em]
\textbf{Proof situation.} The proof wants to rule out proper symplectic resolutions of singular character varieties.\\[0.2em]
\textbf{Reference witness.} A singular \(\mathbb Q\)-factorial terminal variety admits no proper crepant resolution; hence, in the symplectic setting, no proper symplectic resolution.\\[0.2em]
\textbf{Why this case matters.} It separates a general birational obstruction from local corollaries that require extra representation-theoretic hypotheses.
\end{minipage}}

\paragraph{Model behavior.}
\begin{center}
\scriptsize
\renewcommand{\arraystretch}{1.08}
\setlength{\tabcolsep}{3pt}
\resizebox{\linewidth}{!}{%
\begin{tabular}{@{}p{0.18\linewidth}p{0.28\linewidth}ccccccp{0.27\linewidth}@{}}
\toprule
\textbf{Model} & \textbf{Query} & \textbf{Plan} & \textbf{Query} & \textbf{Cite@20} & \textbf{Ground} & \textbf{Suff} & \textbf{Tool} & \textbf{Selected source} \\
\midrule
GPT-5.2 & Q-factorial terminal crepant resolution & \caseN & \caseN & \caseN & \caseY & \caseN & \caseN & Symplectic resolutions and factoriality... \\
Gemini 3.1 Pro & Q-factorial terminal crepant resolution & \caseN & \caseY & \caseN & \caseN & \caseN & \caseN & Global moduli theory of symplectic varieties \\
Claude Opus 4.5 & terminal singularities crepant resolution obstruction & \caseN & \caseY & \caseN & \caseY & \caseY & \caseY & Symplectic resolutions of character varieties \\
DeepSeek V3.2 & character variety singular locus codimension & \caseN & \caseN & \caseN & \caseY & \caseN & \caseN & Symplectic resolutions of character varieties \\
Qwen3-235B Thinking & character variety singular locus codimension & \caseN & \caseN & \caseN & \caseY & \caseN & \caseN & On the codimension of the singular locus \\
Kimi K2 Thinking & character variety singular locus Q-factorial & \caseY & \caseY & \caseN & \caseY & \caseN & \caseN & Symplectic resolutions of character varieties \\
Grok 4 & character variety singular locus codimension & \caseY & \caseY & \caseN & \caseN & \caseN & \caseN & On the codimension of the singular locus \\
\bottomrule
\end{tabular}}
\end{center}

\paragraph{Successful extraction.}
Claude finds a statement equivalent to the general obstruction: a singular \(\mathbb Q\)-factorial terminal variety does not admit a proper crepant resolution, and therefore no symplectic resolution in the relevant setting.

\paragraph{Near misses.}
GPT-5.2, DeepSeek, Qwen, and Kimi ground statements in related sources, but the statements are either specialized to Hamiltonian reductions, codimension estimates, or local facts whose extra hypotheses are not established in the proof context.

\paragraph{Mechanistic diagnosis.}
The proof needs a general birational obstruction: terminal \(\mathbb Q\)-factorial singularities rule out proper crepant, hence symplectic, resolutions.  Several systems instead extract narrower corollaries from the character-variety literature.  These statements are source-supported and locally plausible, but their hypotheses are not supplied by the proof state.  The error is therefore a mismatch in logical portability: the model finds a theorem in the right neighborhood, but not one that can be transported into the current argument.

\paragraph{Takeaway.}
The case shows why sufficiency must be judged relative to the local proof context.  A theorem can be real, relevant, and still unusable if its assumptions encode a more specialized situation than the citing proof has established.

\subsection{Weaker Theorem Under Similar Assumptions: Signless-Laplacian Supersaturation}
\label{app:case_signless_laplacian}

\noindent\fbox{%
\begin{minipage}{0.96\linewidth}
\small
\textbf{Instance.} \texttt{2507.02263v2\_gap\_1}.\quad
\textbf{Paper.} \emph{Some Tur\'{a}n-type results for the signless Laplacian spectral radius}.\quad
\textbf{Domain.} Combinatorics / discrete mathematics.\\[0.2em]
\textbf{Proof situation.} The proof requires a quantitative clique-counting theorem under a signless-Laplacian spectral-radius assumption.\\[0.2em]
\textbf{Reference witness.} If \(k\ge2\), \(\varepsilon>0\), and \(q(G)\ge(1-1/k+\varepsilon)2n\), then \(G\) contains \(\Omega_{k,\varepsilon}(n^{k+1})\) copies of \(K_{k+1}\).\\[0.2em]
\textbf{Why this case matters.} It distinguishes a theorem that produces a large blow-up from the stronger counting statement actually needed by the proof.
\end{minipage}}

\paragraph{Model behavior.}
\begin{center}
\scriptsize
\renewcommand{\arraystretch}{1.08}
\setlength{\tabcolsep}{3pt}
\resizebox{\linewidth}{!}{%
\begin{tabular}{@{}p{0.18\linewidth}p{0.28\linewidth}ccccccp{0.27\linewidth}@{}}
\toprule
\textbf{Model} & \textbf{Query} & \textbf{Plan} & \textbf{Query} & \textbf{Cite@20} & \textbf{Ground} & \textbf{Suff} & \textbf{Tool} & \textbf{Selected source} \\
\midrule
GPT-5.2 & signless Laplacian supersaturation blow-up & \caseY & \caseN & \caseN & \caseY & \caseN & \caseN & Signless Laplacian spectral radius \\
Gemini 3.1 Pro & blow-up graph subgraph containment & \caseN & \caseN & \caseN & \caseN & \caseN & \caseN & Graph containment problems \\
Claude Opus 4.5 & spectral supersaturation complete multipartite subgraph & \caseY & \caseN & \caseN & \caseY & \caseY & \caseY & Signless Laplacian spectral radius \\
DeepSeek V3.2 & signless Laplacian spectral radius complete & \caseN & \caseN & \caseN & \caseNA & \caseNA & \caseNA & none \\
Qwen3-235B Thinking & complete \(k+1\)-partite linear-sized parts & \caseN & \caseN & \caseN & \caseN & \caseN & \caseN & Linear-sized minors... \\
Kimi K2 Thinking & signless Laplacian supersaturation chromatic number & \caseY & \caseY & \caseN & \caseY & \caseY & \caseY & Signless Laplacian spectral radius \\
Grok 4 & blow-up embeddings subgraph counting chromatic & \caseY & \caseN & \caseN & \caseN & \caseN & \caseN & Counting Graph Homomorphisms \\
\bottomrule
\end{tabular}}
\end{center}

\paragraph{Successful extraction.}
Claude and Kimi extract the quantitative counting theorem: spectral excess above the Tur\'{a}n threshold forces \(\Omega(n^{k+1})\) copies of \(K_{k+1}\), which is exactly the scale required downstream.

\paragraph{Near miss.}
GPT-5.2 grounds a true theorem from the same paper family, but it only yields a suitable blow-up with logarithmic-size parts.  That theorem can imply existence of certain subgraphs but not the polynomial number of clique copies needed by the proof.

\paragraph{Mechanistic diagnosis.}
Here the difficulty is not hypothesis mismatch but conclusion strength.  The successful statements and the near miss inhabit the same signless-Laplacian/Tur\'{a}n family, and the assumptions look similar.  The proof, however, needs polynomially many \(K_{k+1}\) copies.  A blow-up or qualitative containment theorem is too weak: it can show existence-like structure without supplying the counting scale used downstream.

\paragraph{Takeaway.}
Assumption checking is not only about preconditions.  It also includes the force of the conclusion: asymptotic order, uniformity, constants, quantifiers, and whether a statement is qualitative or quantitative.  This is a theorem-use skill that ordinary relevance retrieval does not measure.

\subsection{Anchor Consensus, Query Collapse: Muskat Compactness}
\label{app:case_muskat_query_collapse}

\noindent\fbox{%
\begin{minipage}{0.96\linewidth}
\small
\textbf{Instance.} \texttt{1905.05370\_gap\_0}.\quad
\textbf{Paper.} \emph{Weak Solutions to the Muskat Problem with Surface Tension Via Optimal Transport}.\quad
\textbf{Domain.} Analysis / PDE.\\[0.2em]
\textbf{Proof situation.} The proof needs a compactness theorem upgrading uniform bounds and time-translation control into strong \(L^1\) convergence of approximate densities.\\[0.2em]
\textbf{Reference witness.} A convergence theorem for thresholding/variational approximations that gives the strong compactness required to pass to the limit.\\[0.2em]
\textbf{Why this case matters.} It shows that planning can be correct while query generation collapses to generic method words.
\end{minipage}}

\paragraph{Model behavior.}
\begin{center}
\scriptsize
\renewcommand{\arraystretch}{1.08}
\setlength{\tabcolsep}{3pt}
\resizebox{\linewidth}{!}{%
\begin{tabular}{@{}p{0.18\linewidth}p{0.28\linewidth}ccccccp{0.27\linewidth}@{}}
\toprule
\textbf{Model} & \textbf{Query} & \textbf{Plan} & \textbf{Query} & \textbf{Cite@20} & \textbf{Ground} & \textbf{Suff} & \textbf{Tool} & \textbf{Selected source} \\
\midrule
GPT-5.2 & Aubin Lions Wasserstein compactness & \caseY & \caseN & \caseN & \caseNA & \caseN & \caseNA & Nonlinear evolution systems... \\
Gemini 3.1 Pro & Aubin-Lions compactness \(L^1\) convergence & \caseY & \caseN & \caseN & \caseY & \caseN & \caseN & Aubin-Lions-Dubinski\u{i} lemmas \\
Claude Opus 4.5 & Aubin-Lions compactness Wasserstein gradient flow & \caseY & \caseN & \caseN & \caseY & \caseN & \caseN & Fourth-order PDE weak solutions \\
DeepSeek V3.2 & strong \(L^1\) convergence compactness Aubin-Lions & \caseY & \caseN & \caseN & \caseY & \caseN & \caseN & Aubin-Lions-Dubinski\u{i} lemmas \\
Qwen3-235B Thinking & adapted compactness density strong \(L\) & \caseY & \caseN & \caseN & \caseY & \caseN & \caseN & Compactness in adapted weak topologies \\
Kimi K2 Thinking & Aubin-Lions compactness Wasserstein Muskat densities & \caseY & \caseY & \caseN & \caseY & \caseN & \caseN & Fourth-order PDE weak solutions \\
Grok 4 & Aubin-Lions compactness strong \(L^1\) convergence & \caseY & \caseN & \caseN & \caseY & \caseN & \caseN & Aubin-Lions-Dubinski\u{i} lemmas \\
\bottomrule
\end{tabular}}
\end{center}

\paragraph{Mechanistic diagnosis.}
All seven models recognize the broad proof intention, so this is not a planning failure in the usual sense.  The failure occurs when the proof obligation is compressed into search language.  Most systems translate the need into generic ``Aubin--Lions'' or compactness terminology, which retrieves real compactness sources and sometimes grounded statements, but not the theorem used by the Muskat proof.  The lost information is discriminative: the approximation scheme, topology, equation class, and convergence mode that separate the cited tool from a large family of method-level compactness results.

\paragraph{Takeaway.}
The case explains why \(\mathrm{AnchorAcc}\) and query diagnostics cannot substitute for \(\mathrm{ToolAcc}\).  Models can name the right kind of method while failing to preserve the features that make a search query theorem-bearing rather than merely topical.

\subsection{Semantically Plausible Query, Protocol-Sensitive Success: Boolean-to-Tropical Lower Bounds}
\label{app:case_boolean_tropical}

\noindent\fbox{%
\begin{minipage}{0.96\linewidth}
\small
\textbf{Instance.} \texttt{2012.12831\_gap\_0}.\quad
\textbf{Paper.} \emph{Approximation Limitations of Pure Dynamic Programming}.\quad
\textbf{Domain.} Probability / statistics / control.\\[0.2em]
\textbf{Proof situation.} A Boolean lower bound for the assignment/permanent predicate must be instantiated through a Boolean-to-tropical transfer principle.\\[0.2em]
\textbf{Reference witness.} The logical permanent requires monotone Boolean circuits of size at least \(n^{\Omega(\log n)}\).\\[0.2em]
\textbf{Construction citation.} Razborov, \emph{Lower bounds on monotone complexity of the logical permanent}, 1985.
\end{minipage}}

\paragraph{Model behavior.}
\begin{center}
\scriptsize
\renewcommand{\arraystretch}{1.08}
\setlength{\tabcolsep}{3pt}
\resizebox{\linewidth}{!}{%
\begin{tabular}{@{}p{0.18\linewidth}p{0.28\linewidth}ccccccp{0.27\linewidth}@{}}
\toprule
\textbf{Model} & \textbf{Query} & \textbf{Plan} & \textbf{Query} & \textbf{Cite@20} & \textbf{Ground} & \textbf{Suff} & \textbf{Tool} & \textbf{Selected source} \\
\midrule
GPT-5.2 & tropical circuit boolean reduction & \caseN & \caseN & \caseN & \caseY & \caseY & \caseY & Approximation Limitations of Pure Dynamic Programming \\
Gemini 3.1 Pro & Jerrum Snir monotone perfect matching & \caseN & \caseN & \caseN & \caseNA & \caseNA & \caseNA & none \\
Claude Opus 4.5 & tropical circuit monotone boolean simulation & \caseY & \caseY & \caseN & \caseN & \caseN & \caseN & Reciprocal inputs in arithmetic and tropical circuits \\
DeepSeek V3.2 & boolean assignment problem lower bound & \caseY & \caseY & \caseN & \caseNA & \caseNA & \caseNA & none \\
Qwen3-235B Thinking & tropical circuit monotone boolean approximation & \caseY & \caseN & \caseN & \caseY & \caseN & \caseN & Monotone circuit complexity of Boolean functions \\
Kimi K2 Thinking & Jerrum Snir permanent monotone circuit & \caseY & \caseY & \caseN & \caseNA & \caseNA & \caseNA & none \\
Grok 4 & monotone Boolean circuit lower bound & \caseY & \caseY & \caseN & \caseN & \caseN & \caseN & Monotone circuit complexity of Boolean functions \\
\bottomrule
\end{tabular}}
\end{center}

\paragraph{Audit-sensitive outcome.}
The fixed judge scores GPT-5.2 as successful, but the single-annotator audit flags this as an alternative-source false positive: GPT recovers the Boolean-to-tropical transfer theorem from the citing paper, not the monotone lower bound for the logical permanent needed to instantiate the transfer.  This is one of the sensitivity cases reported in Appendix~\ref{app:additional_results}.

\paragraph{Mechanistic diagnosis.}
This case exposes a boundary condition for source admissibility.  The proof contains an internal Boolean-to-tropical transfer principle, but the citation supplies the monotone Boolean lower bound needed to instantiate that principle.  GPT's fixed-judge success retrieves the internal transfer theorem from the citing paper, which is real and proof-relevant but not the external witness.  The audit therefore flags it as a false positive under the external-source protocol.  At the same time, the queries are semantically reasonable, indicating that the failure is not a trivial misunderstanding of the proof architecture.

\paragraph{Takeaway.}
The lesson is that source-grounded evaluation needs an explicit boundary between internal proof machinery and imported tools.  Without that boundary, a model could receive credit for restating the citing paper's own bridge lemma while missing the external theorem that actually supplies the missing premise.

\subsection{Canonical Algebra Query, Zero Retrieval: Tulenbaev Lifting}
\label{app:case_tulenbaev_gluing}

\noindent\fbox{%
\begin{minipage}{0.96\linewidth}
\small
\textbf{Instance.} \texttt{2110.11087v2\_gap\_1}.\quad
\textbf{Paper.} \emph{On the \(\mathbb A^1\)-invariance of \(K_2\) modeled on linear and even orthogonal groups}.\quad
\textbf{Domain.} Algebra / number theory.\\[0.2em]
\textbf{Proof situation.} The proof verifies axioms CFC--HIF for \(K_2(\Phi,-)\).  CFC and DP have been handled; the remaining step uses a Tulenbaev lifting/gluing property for Steinberg groups.\\[0.2em]
\textbf{Reference witness.} For a regular ring \(R\supset k\) and \(\Phi=A_\ell\), \(\ell\ge4\), the Steinberg group \(\mathrm{St}(\Phi,R[t])\) satisfies the polynomial gluing property.\\[0.2em]
\textbf{Construction citation.} Tulenbaev, \emph{The Steinberg group of a polynomial ring}, 1983.
\end{minipage}}

\paragraph{Model behavior.}
\begin{center}
\scriptsize
\renewcommand{\arraystretch}{1.08}
\setlength{\tabcolsep}{3pt}
\resizebox{\linewidth}{!}{%
\begin{tabular}{@{}p{0.18\linewidth}p{0.28\linewidth}ccccccp{0.27\linewidth}@{}}
\toprule
\textbf{Model} & \textbf{Query} & \textbf{Plan} & \textbf{Query} & \textbf{Cite@20} & \textbf{Ground} & \textbf{Suff} & \textbf{Tool} & \textbf{Selected source} \\
\midrule
GPT-5.2 & Tulenbaev lifting Steinberg & \caseN & \caseN & \caseN & \caseNA & \caseNA & \caseNA & none \\
Gemini 3.1 Pro & Steinberg group Tulenbaev lifting property & \caseN & \caseN & \caseN & \caseNA & \caseNA & \caseNA & none \\
Claude Opus 4.5 & Steinberg group Tulenbaev lifting property & \caseY & \caseY & \caseN & \caseNA & \caseNA & \caseNA & none \\
DeepSeek V3.2 & \(K\)-homotopy invariance root system & \caseN & \caseN & \caseN & \caseNA & \caseNA & \caseNA & none \\
Qwen3-235B Thinking & Steinberg groups Tulenbaev lifting simply-laced & \caseN & \caseY & \caseN & \caseNA & \caseNA & \caseNA & none \\
Kimi K2 Thinking & Tulenbaev lifting property Steinberg groups & \caseN & \caseY & \caseN & \caseNA & \caseNA & \caseNA & none \\
Grok 4 & Steinberg groups Tulenbaev lifting property & \caseY & \caseY & \caseN & \caseNA & \caseNA & \caseNA & none \\
\bottomrule
\end{tabular}}
\end{center}

\paragraph{Mechanistic diagnosis.}
This is the cleanest source-acquisition bottleneck in the case set.  Several systems generate queries containing the distinctive author/tool vocabulary, including Tulenbaev, Steinberg groups, and lifting or gluing properties.  Nevertheless, the shared backend does not surface the cited algebra source in the evaluated prefix, and no alternative source is selected.  The model can be mathematically close at the planning and query levels while still failing because the relevant literature is old, terminology varies, or the theorem is poorly indexed.

\paragraph{Takeaway.}
The case motivates theorem-aware retrieval rather than simply larger language models.  A better backend would index statements, assumptions, author/title variants, and theorem-level lexicalizations, so that canonical proof-language queries can reach the relevant source under a reproducible artifact contract.

\subsection{Synthesis}
\label{app:case_synthesis}

Taken together, the cases sharpen the aggregate finding into three distinct failure interfaces.  First, proof obligations must be lexicalized into terms that a scholarly index can surface; the Muskat and Tulenbaev cases show that even good mathematical intent can fail at this interface.  Second, candidate documents must be converted into theorem-bearing passages; the Schoenberg and oracle-rescue cases show that citation recall or source access alone does not solve theorem localization.  Third, sourced statements must be checked against the proof state; the fractional-heat, symplectic-resolution, and signless-Laplacian cases show that true statements can be unusable because of mismatched domains, assumptions, or conclusion strength.  The Brownian case adds the complementary lesson that valid mathematical support need not coincide with the stored citation, while the Boolean-to-tropical case explains why the citing paper itself cannot be used as a shortcut for an external tool.

These examples support the aggregate conclusion in Section~\ref{sec:analysis}.  The weakness is not a single missing module.  Current LLMs can sometimes plan, sometimes retrieve, sometimes ground, and sometimes restate a theorem; what remains unreliable is the disciplined composition of those abilities into a proof-valid source--tool pair.  This is the capability that \(\mathrm{ToolAcc}\) measures and that \ourbench{} is designed to make visible.

\section{Limitations}
\label{app:limitations}

\paragraph{Natural-language evaluation.}
\ourbench{} evaluates source grounding and proof-gap sufficiency in natural-language mathematical documents. This is the right level for the research-assistance setting we target, but it is not a substitute for mechanically checked formal proof. A high \textsc{ToolAcc} score means that the returned statement is source-supported and judged sufficient for the local proof transition; it does not certify a formal derivation. The current validation includes a single-annotator judge-vs-human audit rather than a broad multi-annotator gold set or inter-annotator agreement study. Future releases should add adjudicated expert subsets, borderline-case annotations, and multiple admissible-source labels where feasible.

\paragraph{Retrieval artifact and source access.}
The primary end-to-end protocol uses a shared frozen scholarly-search artifact with Top-20 metadata inspection and at most one source opening. This design makes comparisons reproducible and exposes source-selection pressure, but it also means that failures can arise from model query choice, backend coverage, ranking behavior, source identity matching, or source materialization. We therefore interpret \(\mathrm{CiteRecall@20}\) as an artifact-level diagnostic rather than as a pure measure of model mathematical ability or a complete classical IR evaluation. The same fixed interface also avoids comparing proprietary web-search, PDF parsing, OCR, equation rendering, or multimodal document-reading stacks in the main leaderboard. This improves auditability, but it is stricter than real research workflows and under-evaluates deployed agents that can inspect multiple sources, follow citation chains, or read rendered PDFs. Future tracks should report sensitivity to \(N\), multiple source openings, theorem-aware retrieval backends, and standardized PDF-native or browsing budgets.

\paragraph{Benchmark construction and coverage.}
The current release is built from published, citation-observable proof gaps using a published-first, arXiv-backed pipeline. This favors proof-rich papers with recoverable source text and may underrepresent areas where key results appear primarily in books, older papers, surveys, folklore, publisher-only PDFs, or uncited background knowledge. Eval-200 is also balanced and quality-filtered rather than a random sample of all collected gaps, so the main numbers should be read as controlled diagnostic results on the held-out split rather than as population estimates over all mathematical research. The oracle-source diagnostic has a related boundary: \(\mathrm{OracleCoverage}\) is a release-protocol materialization property, not a model score or a clean upper bound. Non-materialization can arise from ambiguous metadata, unavailable full text, book or survey sources, broken mirrors, or parser failures. The construction audit also shows residual citation-role instrumentality risk: in a minority of audited
retained rows, the author citation is not a clean proof-tool witness, even though the audited
reference-witness sufficiency and external-source checks pass. This limits the interpretation of the
author citation as a historical witness, but not the definition of the official score, which is based on
the final Ground and Suff predicates for the returned source--tool pair.

\paragraph{Scope of the task.}
\ourbench{} intentionally isolates single-gap, source-grounded theorem/tool retrieval. It does not evaluate full autonomous research, theorem discovery, long-horizon proof repair, formalization, or unrestricted browsing. The stored witness is one sufficient route, not a complete characterization of all valid proof strategies; although citation-invariant scoring allows alternative sources, the sufficiency judge may still be conservative when a model closes the gap through a substantially different mathematical route. Because the benchmark is built from published literature, contamination and memorization also cannot be fully ruled out, although source-grounded scoring reduces the value of unsupported memorized statements. The Assisted track is likewise diagnostic rather than primary, since residual anchor-sanitization risk remains; the Raw-track leaderboard does not expose anchors. These boundaries are deliberate: the benchmark narrows the problem enough to make provenance, source selection, and proof-gap sufficiency auditable, while leaving broader browsing, multimodal, and long-horizon agentic settings for future standardized tracks.
\section{Broader Impact}
\label{app:broader_impact}

\ourbench\ is intended to support safer and more useful mathematical research assistance. A system that can retrieve source-grounded tools may help researchers find relevant prior work, avoid missing citations, and check whether a proof step relies on assumptions that are actually available in the literature.

The same capability also has risks. If users over-trust model outputs, an assistant may propagate incorrect attributions, overstate the applicability of a theorem, or present a weakly related source as sufficient. The benchmark therefore emphasizes grounding and sufficiency, and its intended use is as a diagnostic for human-in-the-loop research support rather than as an automatic substitute for scholarly judgment.

There are also access and representation concerns. A retrieval benchmark built from arXiv-backed published papers will naturally favor English-language, openly available, and source-extractable literature. Fixed-budget offline retrieval lowers compute cost relative to unrestricted browsing agents, but it does not remove the broader dependence on search infrastructure and public-index coverage. Future releases should document domain coverage, source accessibility, and any systematic exclusions introduced by the construction pipeline.



\end{document}